\documentclass[10pt,twocolumn,letterpaper]{article}

\usepackage{cvpr}

\usepackage{multirow}
\usepackage[table,xcdraw]{xcolor}
\usepackage{multirow}
\usepackage{graphicx}
\usepackage[table,xcdraw]{xcolor}
\usepackage{pifont}
\usepackage{float}
\usepackage{stfloats}
\usepackage{placeins}
\usepackage{caption}
\usepackage{stackengine}
\usepackage{calc}
\usepackage[export]{adjustbox}
\usepackage{arydshln}
\usepackage{subcaption}
\usepackage{tabularray}
\usepackage{hhline}
\usepackage{bm}
\usepackage{algorithm}
\usepackage{algpseudocode}
\usepackage{amssymb}
\usepackage{wrapfig}
\usepackage{xcolor}
\usepackage{soul}
\usepackage{makecell}
\usepackage{tikz}
\usepackage[normalem]{ulem}
\usepackage{xcolor}
\usepackage{booktabs}
\usepackage[percent]{overpic}

\definecolor{hlblue}{RGB}{220,235,255}
\definecolor{hlgreen}{RGB}{223,245,232}
\definecolor{hlyellow}{RGB}{255,246,205}
\definecolor{pGray}{HTML}{C9CDD3}
\definecolor{pOrange}{HTML}{F4B183}
\definecolor{pLime}{HTML}{A8D08D}
\definecolor{pPurple}{HTML}{B4A7D6}

\UseTblrLibrary{booktabs}

\usepackage[accsupp]{axessibility}

\definecolor{cvprblue}{rgb}{0.21,0.49,0.74}
\usepackage[pagebackref,breaklinks,colorlinks,allcolors=cvprblue]{hyperref}

\def\confName{CVPR}
\def\confYear{2026}

\title{Linear Recurrent Unit with Semantic Modulation for Image Super-Resolution}

\author{
Mingyu Choi\textsuperscript{1}\hspace{3em}
Woo Kyoung Han\textsuperscript{1}\hspace{3em}
Sunghoon Im\textsuperscript{2*}\hspace{3em}
Kyong Hwan Jin\textsuperscript{1*}\\
\textsuperscript{1}Korea University \hspace{3em}
\textsuperscript{2}DGIST\\
{\tt\small \{wookyoung0727, mingyurun, kyong\_jin\}@korea.ac.kr \quad
sunghoonim@dgist.ac.kr}
}

\begin{document}
\maketitle

\begingroup
\renewcommand\thefootnote{}
\footnotetext{*Corresponding author.}
\endgroup

\begin{abstract}
Linear recurrent unit (LRU), designed with a principled formulation for stable linear recurrence, has demonstrated promising accuracy and robustness on long-range dependency tasks. However, its static parameterization and single-scan method limits its applicability to 2D vision tasks. In this study, we propose a LRU-based restoration network with a semantic modulating unit (SMU) to achieve a harmonious balance between performance and efficiency in single-image super-resolution. The SMU plays three key roles: LRU modulation, spatial categorization, and feature enhancement through learned prototype. Extensive experiments demonstrate that our method quantitatively and qualitatively surpasses recent state-of-the-art methods. Notably, our approach achieves superior performance with computational complexity on par with existing methods. The source code and models are available at \href{https://github.com/MingyuChoi-run/LSM}{https://github.com/MingyuChoi-run/LSM}.
\end{abstract}    
\section{Introduction}
\label{sec:introduction}

\begin{figure}
    \centering
    \includegraphics[trim=3 5 7.5 5 ,clip,width = 3.1in]{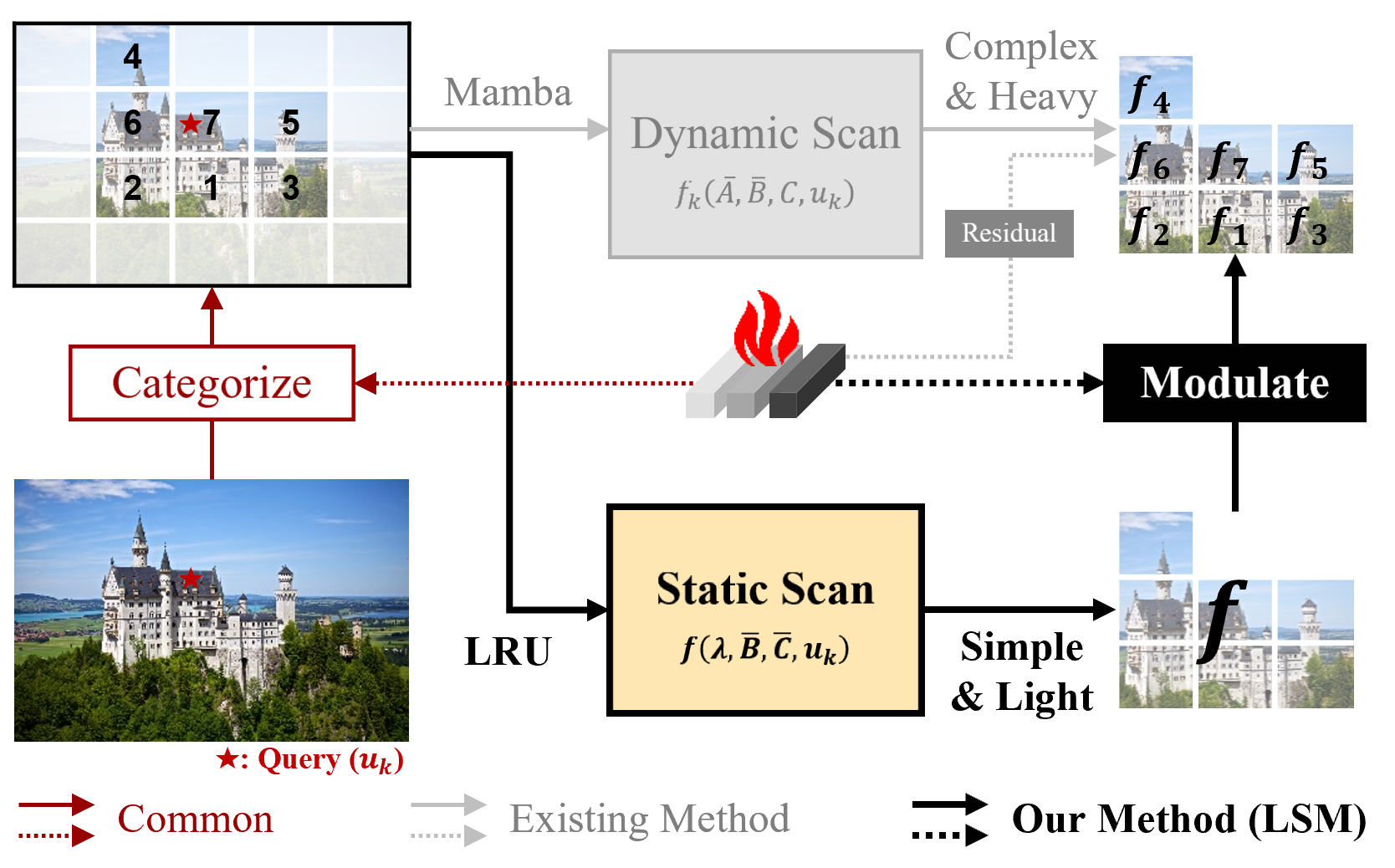}

    \vspace{0pt}

\caption{\textbf{Overall concept of the proposed LSM method.} Unlike conventional complex recurrence in existing method \cite{MambaIRv2}, our method performs a simple and interpretable static scan via LRU \cite{LRU} which is further improved by modulation. Our LSM enables both computational efficiency and semantic-aware enhancement within a single scan.}
    \label{fig:concept}
   \vspace{-15pt}
\end{figure}

Image super-resolution (SR) is a classical ill-posed problem that aims to reconstruct a high-resolution (HR) image from a low-resolution (LR) input. While Transformer-based models \cite{SwinIR, CAT, HAT, RGT, ATD} have achieved high-quality reconstructions, the recent methods with deep state-space models (SSMs), such as Mamba \cite{Mamba}, further advance the line of work \cite{MambaIR, MambaIRv2} with computational efficiency and global receptive fields. However, aforementioned methods employ dynamic parameterization and intricate discretization procedures \cite{S4, S4D} with specialized initializations \cite{HiPPO} that make the model harder to interpret with increased complexity. As illustrated in \cref{fig:concept}, although the existing method \cite{MambaIRv2} has improved the multi-scan property \cite{VMamba, VisionMamba, MambaIR} by categorization, the dynamic scanning of its core architecture limits resource efficiency. 

Recently, to improve resource efficiency, \citet{LRU} proposed linear recurrent units (LRUs), a simplified variant of deep SSMs based on recurrent neural networks (RNNs). LRU achieves linear complexity and stable modeling of long-range sequences through static parameterization, in contrast to SSMs.
Building on the removal of nonlinearity from RNNs that causes inefficiency and  instability \cite{RNNproblem1, RNNproblem2}, LRU \cite{LRU} leverages improved parameterization and initialization techniques based on standard signal propagation logic. Although LRUs have demonstrated strong performance in various long-range sequence modeling \cite{LRA}, their static scanning limits adaptability to complex nonlinear features and spatially varying patterns in the SR task.

To this end, we propose a \textbf{L}inear recurrent unit with \textbf{S}emantic \textbf{M}odulation \textbf{(LSM)}, a novel LRU-based backbone for SR that introduces pixel-wise modulation driven by input-dependent semantics. LSM both preserves the stability of LRUs and enhances adaptivity to spatial context, thereby achieving superior performance while maintaining computational efficiency. 
Our method presented in \cref{fig:concept} is based on lightweight static scanning performed by LRU. We employ a modulating unit with learned dictionary to pre-categorize inputs and modulate the recurrent transitions accordingly. This prototype-conditioned modulation enhances the expressivity of the LRU without incurring the overhead of dynamic scanning.

In summary, our contributions are as follows:
\begin{itemize}
  \item We introduce LSM, the first LRU-based network for SR that unifies linear recurrence and semantic modulation, enabling the reconstruction of high-quality HR images.
  \item We design a semantic modulating unit (SMU) that serves three key roles: 1) modulates the LRU via input-dependent gating, 2) categorizes pixels by semantic similarity to construct more coherent and structured input sequences for the LRU, and 3) enhances feature representations via cross-attention over a learned dictionary.
  \item We demonstrate that the LRU backbone exhibits resource efficiency with respect to input size, which allows allocating more capacity to the SMU and leads to superior performance under limited computational resources.
\end{itemize}
\begin{figure*}
    \centering
    \includegraphics[trim=5 3 6.5 5 ,clip,width=\textwidth]{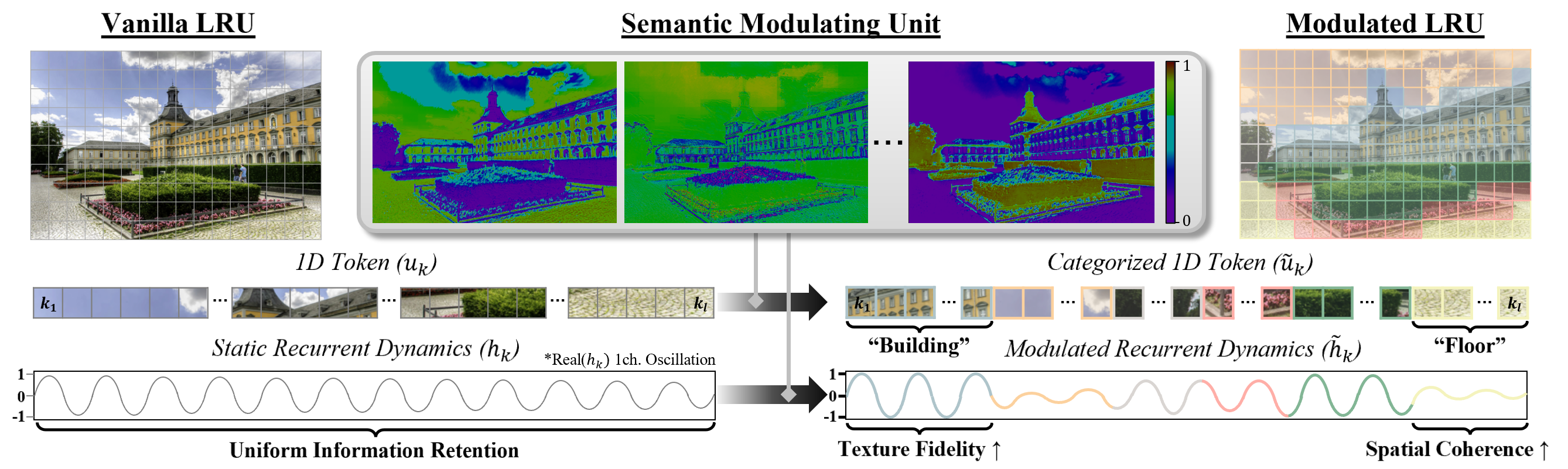}
    \vspace{-17pt}

    \caption{\textbf{Visualization of the modulated LRU.} To extend the static conventional LRU which primarily focuses on preserving information in its hidden state ($h_k$), our method leverages the semantic information from the SMU for SR task. The SMU first categorizes the 1D input token ($u_k$) into $\hat{u}_k$, and then dynamically modulates the hidden state $\hat{h}_k$ to concentrate on critical information according to its category, thereby endowing it with dynamic modulation characteristics.}
    \label{fig:method}
    \vspace{-15pt}
\end{figure*}
\section{Related Works}
\label{sec:relatedworks}

\noindent \textbf{Image Super-Resolution}
Since the advent of deep learning, SR has progressed rapidly.  Early CNN-based methods such as SRCNN \cite{SRCNN} demonstrated that a compact convolutional stack could surpass classical approaches \cite{VDSR, EDSR, LTE}. Subsequent work deepened networks to enhance representational capacity: VDSR \cite{VDSR} introduced residual learning, EDSR \cite{EDSR} simplified the residual block to enable deeper models, and RDN \cite{RDN} exploited dense connections.
To overcome the limited receptive field, attention mechanisms \cite{Transformer} were incorporated into CNN architectures, yielding notable improvements. Various attention methods such as channel attention \cite{RCAN}, secondary attention \cite{SAN}, global attention \cite{HAN}, and non-local sparse attention \cite{NLSA} have been proposed. CNN-attention hybrids extended the modeling capacity of local convolution operators.
More recently, Transformer-based models \cite{ViT, SwinT} have leveraged self-attention to capture long-range dependencies. 
Patch-based approaches \cite{IPT}, local shifted windows \cite{SwinIR}, sparse attention \cite{ART, OmniSR}, multi-scale methods \cite{ELAN}, and anchored attention \cite{GRL} have been proposed to mitigate computational complexity and improve scalability to HR inputs.
Further works \cite{CAT, DAT, RGT} explore efficient global spatial interaction through various aggregation mechanisms across local windows. 
To improve adaptability and efficiency of attentions, PromptIR \cite{PromptIR} adopts task-aware prompting, whereas ATD \cite{ATD} utilizes dictionary-driven priors.
Recently, since deep SSMs \cite{S4, S4D, S5, Mamba} have emerged as powerful tools for modeling long input sequences, Mamba-based \cite{Mamba} models have been actively expanding their applications to vision tasks. VMamba \cite{VMamba} and VisionMamba \cite{VisionMamba} applied multi-scan processing tailored to the image domain, while MambaIR \cite{MambaIR} and MambaIRv2 \cite{MambaIRv2} extended this to low-level image restoration. Drawing inspiration from previous studies, we design our LSM to learn SMU that improves the computational efficiency of SSMs and enhances adaptivity.

\noindent \textbf{Linear Recurrent Unit}
Traditionally, RNNs \cite{RNN1, RNN2, RNN3} have long been used to model sequential dependencies. However, they suffer from gradient instability \cite{RNNproblem1, RNNproblem2}. Various studies have been conducted to overcome this \cite{RNNimprove1, GRU, RNNimprove2}, and recent research has revisited their connection to SSMs \cite{LRU}. With principled parameterization and normalization, deep RNNs achieve performance comparable to continuous state-space formulations.
The linear recurrent unit (LRU) \cite{LRU} embodies this idea, providing stable and expressive linear recurrence while maintaining efficiency. The real-gated LRU (RG-LRU) \cite{RGLRU}, gated variants of LRU, incorporates input-dependent modulation \cite{LSTM, GRU} to further enhance expressivity. RG-LRU was further extended to the large language model domain through \emph{Griffin}, a hybrid architecture that combines gated linear recurrences with local attention. This model achieves competitive performance with strong baselines \cite{Mamba} while using significantly fewer tokens. Our method extends RG-LRU \cite{RGLRU} by applying LRU \cite{LRU} to spatial sequences in SR and incorporating a lightweight modulation mechanism that adapts the recurrence behavior to pixel-wise semantics.
\section{Methodology}
\label{sec:methodology}

\begin{figure*}
    \centering
    \includegraphics[trim=0 1 0 1 ,clip,width=\textwidth]{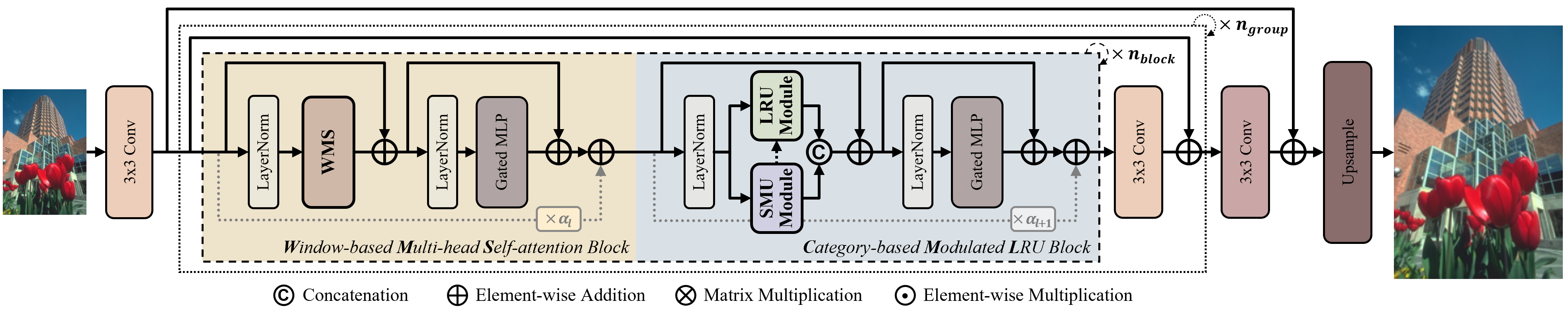}

    \vspace{-10pt}

    \caption{\textbf{Overall Architecture of LSM.} The LSM features a sequential and recursive structure, similar to a RG-LRU. It comprises a preceding Window-based Multi-head Self-attention (WMS) block and a subsequent Category-based Modulated LRU (CML) block. Each block incorporates a Gated MLP and a scaled skip connection $\alpha$. The CML effectively operates through the synergistic interaction between the LRU and SMU modules.}
    \label{fig:main}
    \vspace{-13pt}
\end{figure*}

\subsection{Preliminaries}
LRU \cite{LRU} is designed to overcome the inefficiencies of conventional RNNs by enabling stable, parallelizable, and expressive recurrent computation. To arrive at the efficient and normalized diagonal recurrence used in our model, we follow the canonical LRU formulation. The complete derivation from standard RNNs to LRUs involves a sequence of transformations including linearization of recurrence, complex diagonalization, exponential eigenvalue parameterization, and normalization. These derivation steps are detailed in the \textit{supplementary material}. The final LRU update used in our method is as follows:
\begin{equation}
\begin{split}
\bar{h}_k &= \operatorname{diag}(\lambda) \bar{h}_{k-1} + \gamma \odot (\bar{B}\, u_k), \\
\bar{y}_k &= \bar{C}\, \bar{h}_k + D u_k,
\label{eq:lru}
\end{split}\tag{1}
\end{equation}
\noindent where $u_k$ is the input vector at step $k$, $\bar{h}_k$ is the hidden state, $\lambda$ is an eigenvalue which is reparameterized with magnitude $\nu$ and phase $\theta$ as $\lambda_j = \exp\!\big(-\exp(\nu^{\log}_j)\big) \exp\!\big(i\, \exp(\theta^{\log}_j)\big)$. A normalization factor $\gamma$ is  defined as $\gamma_j = \sqrt{1 - |\lambda_j|^2}$, and $\odot$ denotes element-wise multiplication.

\subsection{Motivation}
Although LRU offers a stable, interpretable, and efficient core, a static linear recurrence used as a global block is limiting for 2D vision tasks. LRU's limitation is reflected in the initialization behavior shown in \cref{fig:method}. In vanilla LRU, the recurrence dynamics remain identical across all tokens, as $\lambda$, $B$, and $C$ are static in both space and time. These static characteristics impose significant constraints on low-level visual tasks that must individually capture distinct spatial features. In particular, both long-range contextual information and detailed local features are crucial in SR task. The expressive limitations of applying vanilla LRU globally are demonstrated in the inference results presented in Section~\ref{sec:ablation}. Consequently, despite its success in sequential data modeling, LRU has not been widely adopted as a backbone in the image domain.
As discussed in Section~\ref{sec:relatedworks}, RG-LRU \cite{RGLRU} and its instantiation in \emph{Griffin} address the limitations of static recurrence by introducing input-dependent gating mechanisms. The core of \emph{Griffin} is summarized as follows: \\
\begin{equation}
\begin{split}
r_k = \sigma(W_a u_k + b_a), \\
i_k = \sigma(W_x u_k + b_x), \\
a_k = a^{c r_k},
\end{split}\tag{2}
\label{eq:griffin}
\end{equation}
\noindent where $r_k$ is the \textit{recurrence gate}, $i_k$ is the \textit{input gate}, and $a_k$ is the \textit{recurrent weight}. Note that $\sigma$ denotes the sigmoid function. We parameterize $a$ in \cref{eq:griffin} as $a=\sigma(\Lambda_{\mathrm{RG}})$, where $\Lambda_{\mathrm{RG}}$ is a learnable parameter that guarantees $0\le a\le 1$. Here, $c>0$ is a constant that controls the sharpness of $a_k = a^{c r_k}$. The hidden-state update then can be presented as follows:
\begin{equation}
\begin{split}
h_k &= a_k \odot h_{k-1} + \sqrt{1-a_k^2}\, \odot (i_k \odot u_k).
\end{split}\tag{3}
\label{eq:rglru}
\end{equation}
$r_k$ dynamically adjusts $a_k$ based on $u_k$, allowing flexible determination of how strongly past information is retained at specific sequence points. $r_k$ helps reduce the influence of irrelevant inputs and preserve crucial information over long durations. $i_k$ dynamically controls how much of $u_k$ is integrated into the hidden state. $i_k$ provides the ability to discern the importance of incoming information, filtering noisy or irrelevant inputs and focusing on meaningful ones to update the new hidden state. Furthermore, the serial deployment strategy of \emph{Griffin} combines the ability of RG-LRU to learn long-range dependencies with the capacity of local attention for precise local context modeling, creating synergistic effects.

\begin{figure}[t]
    \centering
    \includegraphics[trim=0.5 5 3 3 ,clip,width = 3.3in]{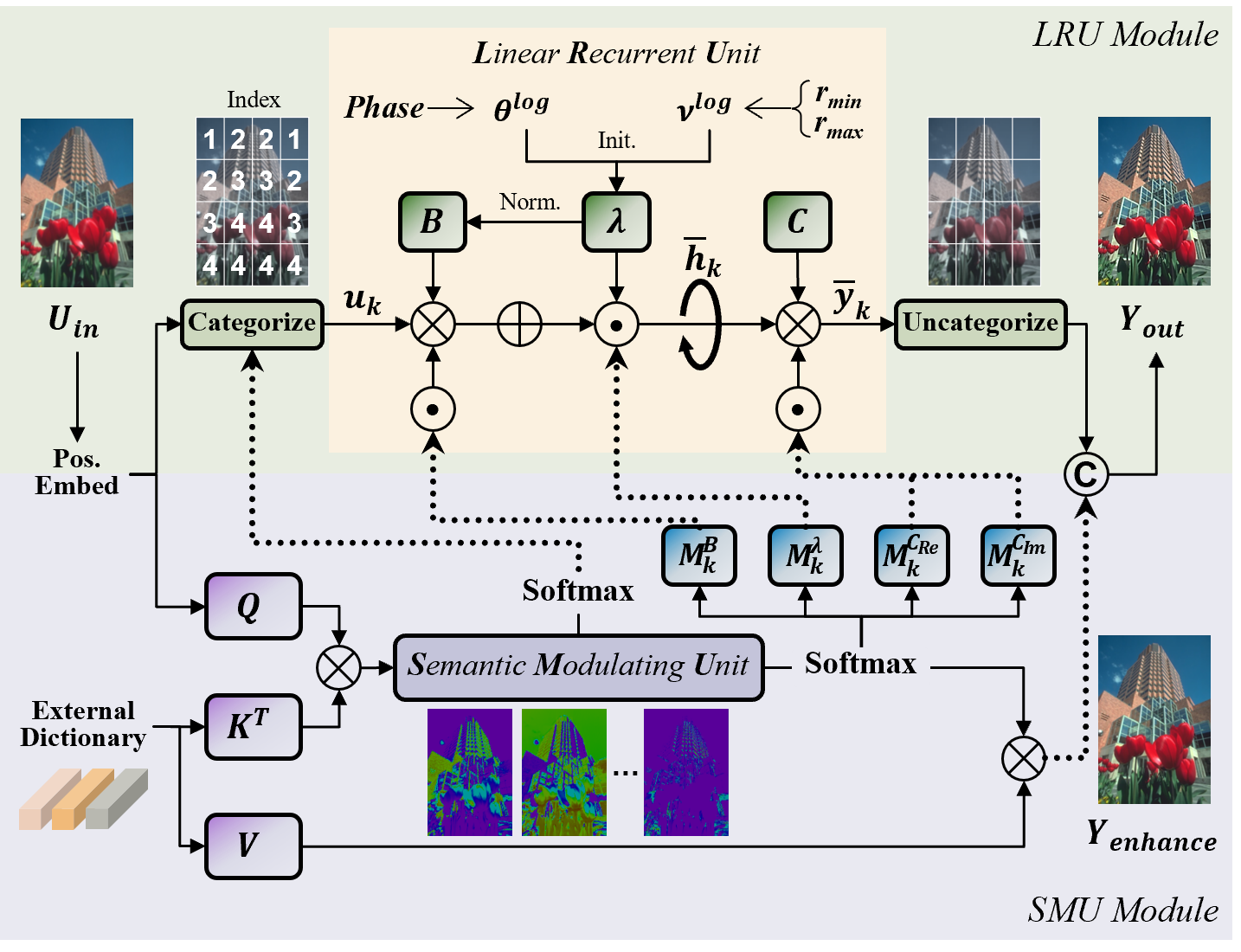}
    \vspace{-15pt}
\caption{\textbf{Proposed LRU and SMU.} The LRU initializes ($\lambda$) and normalizes ($B$) its state transition parameters. The SMU generated by operations between feature ($Q$) and dictionary ($K$) simultaneously categorizes the input injecting LRU ($u_k$), dynamically modulates its transition matrices via modulating tokens ($\mathbf{M}_k$), and enhances features through cross-attention.}
    \label{fig:network}
   \vspace{-15pt}
\end{figure}

\subsection{Category-based Modulated LRU}
As shown in \cref{fig:network}, the semantic modulating unit (SMU) enhances the expressivity and adaptivity required for SR, drawing insights from RG-LRU and its core equations. Leveraging similarity with a dictionary \cite{DictionaryLearning}, it performs three key roles. These roles are complementary and contribute to the performance of LSM. We describe each component and its function in the following.

\noindent \textbf{LRU Modulation}
We first conduct a thorough analysis of the RG-LRU \cite{RGLRU} structure from the perspective of vanilla LRU \cite{LRU} to gain new insights into incorporating data-dependent modulation. Revisiting the hidden state update in \cref{eq:rglru}, the dynamic modulation of $a_k$ controlled by $r_k$ can be interpreted as an input-adaptive recurrent coefficient, a mechanism absent in the fixed $\lambda$ of vanilla LRU. Building on this insight, we model the hidden state update \cref{eq:rglru} as:
\begin{equation}
\begin{split}
h_k &= a^{c r_k} \odot h_{k-1} + \sqrt{1-(a^{c r_k})^2}\, \odot (i_k \odot u_k), \\
&\approx (a^c \odot F_1(r_k)) \odot h_{k-1} \\
&+(\sqrt{1-(a^{c})^2} \odot F_2(r_k) \odot i_k) \odot u_k,
\end{split}\tag{4}
\label{eq:approx_rglru}
\end{equation}
where $r_k$, $i_k$ are input-dependent gates. The first term $a^{c r_k}$ represents a non-linear modulation of the recurrent weight $a$ by $r_k$ and constant $c$, intuitively approximated as a non-linear gate $F_1(r_k)$ multiplied by $a^c$, which is applied to $h_{k-1}$. For the second term, $\sqrt{1-(a^{c r_k})^2}$ dynamically changes between $0$ and $\sqrt{1-a^{2c}}$ depending on $r_k$. Thus, combined with $i_k$, this term can be approximated as an input scaling factor $\gamma$ multiplied by a non-linear gate $F_2(r_k)$ and $i_k$, which dynamically scales $u_k$.
Guided by \cref{eq:approx_rglru}, we augment \cref{eq:lru} by incorporating vision-specific input-dependent modulation into the hidden state update, aiming to enhance its expressivity:
\begin{equation}
\begin{split}
\bar{h}_k &= (\lambda \odot M_k^{\lambda}) \odot \bar{h}_{k-1} + \gamma \odot (\bar{B}\, u_k) \odot M_k^{B}, \\
\bar{y}_k &= \bar{C}\, \bar{h}_{k} \odot M^{C}_{k} + \bar{D}\, u_k,
\end{split}\tag{5}
\label{eq:LSM}
\end{equation}
where $M_k^{\lambda}$ and $M_k^B$ are modulating tokens generated from a separate network dependent on $u_k$. 
Furthermore, to enhance output expressivity, $M_k^C$ is applied to the complex-valued $\bar{C}$. It is decomposed into $M_k^{C_{\mathrm{Re}}}$ and $M_k^{C_{\mathrm{Im}}}$ to modulate the real and imaginary components of $\bar{C}$, respectively. This leads to the proposed modulated LRU output equation.

\noindent \textbf{Modulating Tokens}
To generate modulating tokens, we use the SMU. A dictionary $\mathbf{D} \in \mathbb{R}^{P \times c}$ serves as a compact set of learned prototype tokens \cite{ATD} with $P \ll T$, where $T$ is the number of image tokens. Given input features $U \in \mathbb{R}^{T \times c}$ which is the output of the local attention block, we apply separate linear projections for numerical stability and compute a cosine-similarity map with temperature $\tau$:
\begin{equation}
\begin{split}
Q_U &= \operatorname{Linear}_Q(U), \\
K_{\mathbf{D}},\, V_{\mathbf{D}} &= \operatorname{Linear}_K(\mathbf{D}),\, \operatorname{Linear}_V(\mathbf{D}), \\
\text{SMU} &= \operatorname{Sim_{cos}}(Q_U, K_{\mathbf{D}})/\tau,
\end{split}\tag{6}
\end{equation}
with $Q_U \in \mathbb{R}^{T \times c/3}$, $K_{\mathbf{D}} \in \mathbb{R}^{P \times c/3}$, and $V_{\mathbf{D}} \in \mathbb{R}^{P \times c/2}$. The modulating tokens are then derived by chunking a softmax-normalized affinity:
\begin{equation}
\begin{split}
\mathbf{M}_k &= \operatorname{Chunk}\!\big(\operatorname{SoftMax}(\text{SMU}),\,4\big),
\end{split}\tag{7}
\end{equation}
where $\mathbf{M}_k$ represents the vector of $M_k^{\lambda}$, $M_k^{B}$, $M_k^{C_{\mathrm{Re}}}$, and $M_k^{C_{\mathrm{Im}}}$. This enables pixel-wise modulation as \cref{eq:LSM} with minimal overhead, relying solely on the chunking method.\\
\noindent \textbf{Multi-role of SMU}
The SMU additionally serves two roles beyond parameter modulation. We empirically demonstrate these roles in \cref{sec:ablation}: \\
\textbf{(i) Semantic categorization to complement single-scan}
Standard 1D recurrence \cite{RNN1, LRU} struggles to relate spatially distant yet semantically similar pixels. To address this, SMU assigns each pixel to one of $P$ semantic groups using a temperature-controlled Gumbel softmax \cite{Gumbel}, followed by $\operatorname{argmax}$ for hard assignment. Pixels are then categorized accordingly before being processed by the modulated LRU, allowing semantically related but distant pixels to interact within a single scan. \\
\textbf{(ii) Semantic-aware global feature enhancement}
In parallel, the affinity $\operatorname{SoftMax}(\text{SMU})$ performs attention over the value embeddings as $\operatorname{SoftMax}(\text{SMU}) \cdot V_{\mathbf{D}}$, enabling aggregation of globally relevant information across the entire spatial domain. This results in a better representation of features $Y_{\text{enhance}} \in \mathbb{R}^{T \times c/2}$ which is concatenated with the output of LRU $Y_{\text{LRU}} \in \mathbb{R}^{T \times c/2}$. The final output is obtained as
$Y_{\text{out}} = \operatorname{Concat}(Y_{\text{LRU}}, Y_{\text{enhance}})$. This pathway provides complementary global context beyond the single-scan recurrence. 

\begin{figure}[t]
\footnotesize
\centering
\hspace{-18pt}

\hspace{-1pt}
{\includegraphics[trim=5 5 -25 7.2,clip,width = 1.05in]{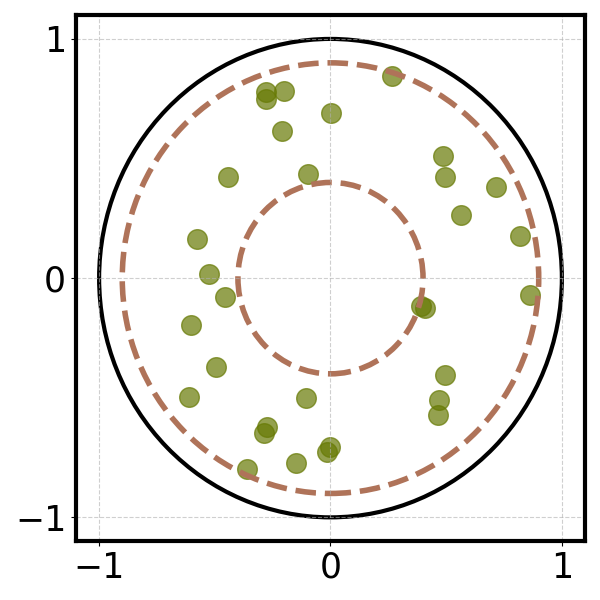}}
\hspace{-1pt}%
{\includegraphics[trim=5 5 -20 7,clip,width = 1.04in]{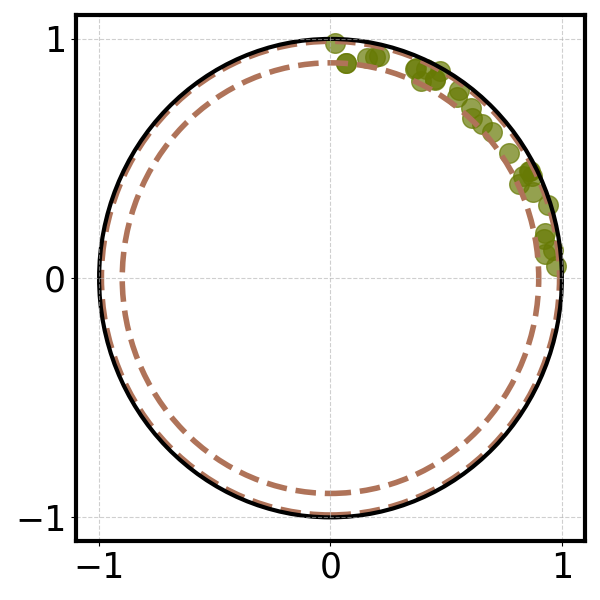}}
\hspace{-1pt}%
{\includegraphics[trim=5 5 -20 7.3,clip,width = 1.04in]{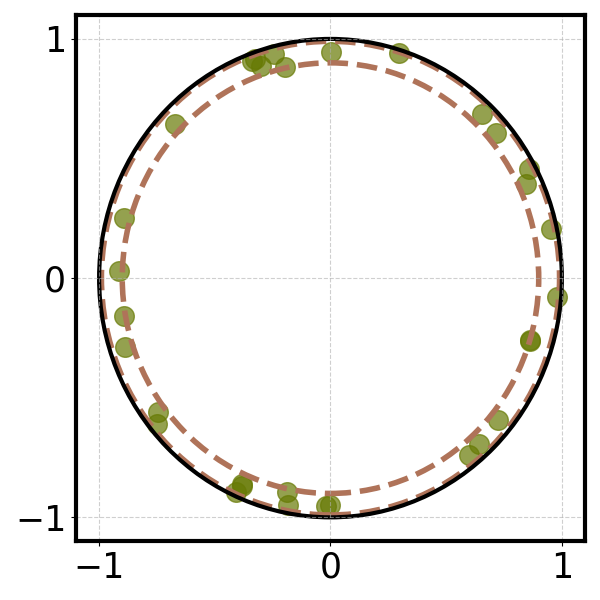}}

\hspace{0pt}
\stackunder[2pt]{\includegraphics[trim=0 0 0 0,clip,width = 1.05in]{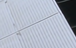}}{$0.4 / 0.9 / 2\pi$}
\hspace{0pt}%
\stackunder[2pt]{\includegraphics[trim=0 0 0 0,clip,width = 1.05in]{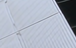}}{$0.9 / 0.99 / 0.5\pi$}
\hspace{0pt}%
\stackunder[2pt]{\includegraphics[trim=0 0 0 0,clip,width = 1.05in]{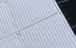}}{$0.9 / 0.99 / 2\pi$}

\vspace*{-5pt}
\caption{Qualitative ablation study of $\lambda$ initialization. Ringing initialization of eigenvalues in the complex plane with varying $r_{min}/r_{max}/\theta_{max}$ (top), and the corresponding SR images on Urban100 dataset (bottom).}
\label{fig:ablation}
\vspace*{-15pt}
\end{figure}

\subsection{The Overall Network Architecture}
 We follow design of \textit{Griffin} \cite{RGLRU} but merge the global attention and recurrent functions into a single category-based modulated LRU (CML) block. This integration forms the key distinction of LSM. Thus, the network consists of two serial stages: a local block and a global block. As shown in \cref{fig:main}, a $3\times3$ convolution first extracts shallow features, followed by a $l^{th}$ local window-based multi-head self-attention (WMS) block and then the $(l+1)^{th}$ CML block. To combine the distinct self-attention and LRU representations, an adaptive residual skip with learnable weight $\alpha$ is employed \cite{RGT}. Each block follows the Transformer pattern of layernorm \cite {Layernorm}, token mixing layer, layernorm, and gated MLP \cite{Transformer}, where token mixing is replaced by MHSA \cite{SwinIR, CAT} or modulated LRU. In detail, the token mixing in the CML block consists of the LRU and SMU modules. As illustrated in \cref{fig:network}, the SMU injects key contextual cues into the LRU via a trainable dictionary. When $l = 0, 2, 4, \dots$, the local–global block pairs are repeated until the maximum index $n_{\text{block}}$, forming one hierarchical group. These groups are then stacked $n_{\text{group}}$ times, and finally upsampled with pixel-shuffle \cite{PixelShuffle} to reconstruct the HR output.

\section{Experiments}
\label{sec:experiments}
\subsection{Experimental Settings}
In classic SR task, we provide two model versions, LSM-S and LSM with different computational complexity. LSM-S uses 6 groups, each with 5 blocks and 180 channels. In the CML block, the LRU state size is 32. For initializing $\lambda$, we set $r_{\min}=0.9$, $r_{\max}=0.99$, and $\theta_{max}=2\pi$. Since the learned dictionary has 128 categories, the SMU outputs a 128 channel vector. We then apply a channel-wise split into four 32 channel modulating tokens, and each token is fed into the LRU via element-wise multiplication. For LSM, we increase $n_{\text{group}}$ from 6 to 8 and all other settings match LSM-S. 
Furthermore, we also provide a lightweight version of our model, named LSM-light. In LSM-light, the $n_{\text{group}}$ is reduced to 4, with each group consisting of 6 blocks and 60 channels. The number of LRU states and categories in the dictionary is also halved to 16 and 64, respectively. More training details are provided in the \textit{supplementary material}.

\subsection{Ablation Study}
\label{sec:ablation}
We conduct ablation experiments to validate the proposed design. All models are trained on DIV2K \cite{DIV2K} and Flickr2K \cite{EDSR} under identical $\times2$ settings for 200k iterations, and evaluated on Set14 \cite{Set14}, Manga109 \cite{Manga109}, and Urban100 \cite{Urban100} benchmarks.
\begin{table}[]
\captionsetup{justification=centering}
\centering
\footnotesize
\caption{Quantitative ablation study of $\lambda$ initialization.}
\vspace{-5pt}
\resizebox{\columnwidth}{!}{%
\begin{tabular}{ccc|cc|cc|cc}
\hline
\multirow{2}{*}{\bm{$r_{min}$}} & \multirow{2}{*}{\bm{$r_{max}$}} & \multirow{2}{*}{\bm{$\theta_{max}$}} & \multicolumn{2}{c|}{\textbf{Set14}} & \multicolumn{2}{c|}{\textbf{Urban100}} & \multicolumn{2}{c}{\textbf{Manga109}} \\
                               &                                &                                 & \textbf{PSNR}    & \textbf{SSIM}   & \textbf{PSNR}     & \textbf{SSIM}     & \textbf{PSNR}     & \textbf{SSIM}     \\
\hline
0.4                            & 0.9                            & $2\pi$                              & 34.37            & 0.9242          & 33.96             & 0.9434            & 39.91             & 0.9797            \\
0.9                            & 0.99                           & $0.5\pi$                             & 34.45            & 0.9250           & 33.95             & 0.9431            & 39.89             & 0.9797            \\
0.9                            & 0.99                           & $2\pi$                              & \textbf{34.56}            & \textbf{0.9255}          & \textbf{34.04}             & \textbf{0.9435}            & \textbf{40.01}             & \textbf{0.9802}           \\
\hline
\end{tabular}%
}
\label{tab:initial}
\vspace{-6pt}
\end{table}
\begin{table}[]
\captionsetup{justification=centering}
\centering
\caption{Ablation study on the multi-role of the SMU}
\vspace{-5pt}
\resizebox{\columnwidth}{!}{%
\begin{tabular}{cccc|c|cc|cc}
\hline
\multirow{2}{*}{\textbf{LRU}} & \multirow{2}{*}{\textbf{Categorize}} & \multirow{2}{*}{\textbf{CrossAttn}} & \multirow{2}{*}{$\mathbf{M}_k$} & \multirow{2}{*}{\textbf{\# Params}} & \multicolumn{2}{c|}{\textbf{Set14}} & \multicolumn{2}{c}{\textbf{Manga109}} \\
                              &                                     &                              &                                &                                     & \textbf{PSNR}  & \textbf{SSIM} & \textbf{PSNR}   & \textbf{SSIM}   \\
\hline
\ding{51}                             &                                     &                              &                                & \textbf{9.06M}                                & 34.44              & 0.9246        & 39.93               & 0.9797          \\
\ding{51}                             & \ding{51}                                   &                              &                                & 9.33M                                & 34.42              & 0.9248        & 39.96               & 0.9799          \\
\ding{51}                             & \ding{51}                                   & \ding{51}                            &                                & 9.72M                                & 34.46              & 0.9252        & 39.95               & 0.9800            \\
\ding{51}                             & \ding{51}                                   & \ding{51}                            & \ding{51}                              & 9.72M                                & \textbf{34.56}              & \textbf{0.9255}        & \textbf{40.01}               & \textbf{0.9802}         \\
\hline
\end{tabular}%
}
\label{tab:multirole}
\vspace{-6pt}
\end{table}
\begin{table}[]
\captionsetup{justification=centering}
\centering
\setlength{\tabcolsep}{1.2pt}
\footnotesize
\caption{Ablation study on the effectiveness of modulating tokens}
\vspace{-5pt}
\resizebox{\columnwidth}{!}{%
\begin{tabular}{ccccc|c|cc|cc|cc}
\hline
\multirow{2}{*}{$M_k^\lambda$} & \multirow{2}{*}{$M_k^B$} & \multirow{2}{*}{$M_k^{\mathrm{C_{Re}}}$} & \multirow{2}{*}{$M_k^{\mathrm{C_{Im}}}$} & \multirow{2}{*}{\textbf{Linear}} & \multirow{2}{*}{\textbf{\# Params}} & \multicolumn{2}{c|}{\textbf{Set14}} & \multicolumn{2}{c|}{\textbf{Urban100}} & \multicolumn{2}{c}{\textbf{Manga109}} \\
                            &                             &                               &                               &                                  &                                     & \textbf{PSNR}    & \textbf{SSIM}   & \textbf{PSNR}     & \textbf{SSIM}     & \textbf{PSNR}     & \textbf{SSIM}     \\
\hline
                            &                             &                               & \multicolumn{1}{l}{}          &                                  & \textbf{9.72M}                                & 34.46            & 0.9252          & 33.92             & 0.9432            & 39.95             & 0.9800            \\
                            &                             & \ding{51}                             & \ding{51}                             &                                  & \textbf{9.72M}                                & 34.42            & 0.9248          & 34.00             & \textbf{0.9436}            & 39.94             & 0.9799            \\
\ding{51}                           & \ding{51}                           & \ding{51}                             & \ding{51}                             &                                  & \textbf{9.72M}                                & 34.56            & 0.9255          & \textbf{34.04}             & 0.9435            & \textbf{40.01}             & \textbf{0.9802}            \\
\ding{51}                           & \ding{51}                           & \ding{51}                             & \ding{51}                             & \ding{51}                                & 9.91M                                & \textbf{34.60}            & \textbf{0.9259}          & 33.97             & 0.9434            & 39.91             & 0.9798           \\
\hline
\end{tabular}%
}
\label{tab:M_t}
\vspace{-13pt}
\end{table}
\setlength\dashlinedash{0.5pt}
\setlength\dashlinegap{1pt}
\begin{table*}[!t]
\captionsetup{justification=centering}
\centering
\footnotesize
\caption{Quantitative comparison on classic SR with state-of-the-art methods. The best and second best results are in {\color[HTML]{FF0000}red} and {\color[HTML]{0000FF}blue}.}
\vspace{-5pt}
\resizebox{\textwidth}{!}{%
\begin{tabular}{l|c|c|cc|cc|cc|cc|cc}
\hline
                         &                         &                             & \multicolumn{2}{c|}{Set5}                                     & \multicolumn{2}{c|}{Set14}                                    & \multicolumn{2}{c|}{B100}                                     & \multicolumn{2}{c|}{Urban100}                                 & \multicolumn{2}{c}{Manga109}                                 \\
\multirow{-2}{*}{Method} & \multirow{-2}{*}{Scale} & \multirow{-2}{*}{\# Params} & PSNR                         & SSIM                          & PSNR                         & SSIM                          & PSNR                         & SSIM                          & PSNR                         & SSIM                          & PSNR                         & SSIM                          \\
\hline
EDSR \cite{EDSR}                     & $\times2$                      & 42.6M                       & 38.11                        & 0.9602                        & 33.92                        & 0.9195                        & 32.32                        & 0.9013                        & 32.93                        & 0.9351                        & 39.10                        & 0.9773                        \\
\hdashline
SwinIR \cite{SwinIR}                   & $\times2$                      & 11.8M                       & 38.42                         & 0.9623                        & 34.46                        & 0.9250                        & 32.53                        & 0.9041                        & 33.81                        & 0.9427                        & 39.92                        & 0.9797                        \\
CAT-A \cite{CAT}                    & $\times2$                      & 16.5M                       & 38.51                        & 0.9626                        & 34.78                        & 0.9265                        & 32.59                        & 0.9047                        & 34.26                        & 0.9440                        & 40.10                        & 0.9805                        \\
DAT-S \cite{DAT}                    & $\times2$                      & 11.1M                       & 38.54                        & 0.9627                        & 34.60                        & 0.9258                        & 32.57                        & 0.9047                        & 34.12                        & 0.9444                        & 40.17                        & 0.9804                        \\
ART \cite{ART}                      & $\times2$                      & 16.4M                       & 38.56                        & 0.9629                        & 34.59                        & 0.9267                        & 32.58                        & 0.9048                        & 34.30                        & 0.9452                        & 40.24                        & 0.9808                        \\
HAT-S \cite{HAT}                    & $\times2$                      & 9.5M                       & 38.58                        & 0.9628                        & 34.70                        & 0.9261                        & 32.59                        & 0.9050                        & 34.31                        & 0.9459                        & 40.14                        & 0.9805                        \\
RGT-S \cite{RGT}                    & $\times2$                      & 10.1M                       & 38.56                        & 0.9627                        & 34.77                        & 0.9270                        & 32.59                        & 0.9050                        & 34.32                        & 0.9457                        & 40.18                        & 0.9805                        \\
\hdashline
MambaIR \cite{MambaIR}                  & $\times2$                      & 20.4M                       & 38.57                        & 0.9627                        & 34.67                        & 0.9261                        & 32.58                        & 0.9048                        & 34.15                        & 0.9446                        & 40.28                        & 0.9806                        \\
MambaIRv2-S \cite{MambaIRv2}              & $\times2$                      & 9.6M                        & 38.53                        & 0.9627                        & 34.62                        & 0.9256                        & 32.59                        & 0.9048                        & 34.24                        & 0.9454                        & 40.27                        & 0.9808                        \\
\hdashline
\rowcolor{gray!10} LSM-S (Ours)                   & $\times2$                      & 9.7M                        & 38.60                        & 0.9628                        & 34.66                        & 0.9264                        & 32.60                        & {\color[HTML]{0000FF} 0.9051}                        & 34.40                        & 0.9464                        & 40.25                        & 0.9805                        \\
\rowcolor{gray!10} LSM (Ours)                   & $\times2$                      & 12.8M                       & {\color[HTML]{0000FF} 38.62} & {\color[HTML]{0000FF} 0.9630} & {\color[HTML]{0000FF} 34.82} & {\color[HTML]{FF0000} 0.9272} & {\color[HTML]{0000FF} 32.61} & {\color[HTML]{0000FF} 0.9051} & {\color[HTML]{0000FF} 34.43} & {\color[HTML]{0000FF} 0.9466} & {\color[HTML]{0000FF} 40.35} & {\color[HTML]{0000FF} 0.9809} \\
\rowcolor{gray!10} LSM+ (Ours)                  & $\times2$                      & 12.8M                       & {\color[HTML]{FF0000} 38.66} & {\color[HTML]{FF0000} 0.9631} & {\color[HTML]{FF0000} 34.83} & {\color[HTML]{0000FF} 0.9271} & {\color[HTML]{FF0000} 32.64} & {\color[HTML]{FF0000} 0.9054} & {\color[HTML]{FF0000} 34.61} & {\color[HTML]{FF0000} 0.9475} & {\color[HTML]{FF0000} 40.47} & {\color[HTML]{FF0000} 0.9812} \\
\hline
EDSR \cite{EDSR}                     & $\times3$                      & 43.0M                       & 34.65                        & 0.9280                        & 30.52                        & 0.8462                        & 29.25                        & 0.8093                        & 28.80                        & 0.8653                        & 34.17                        & 0.9476                        \\
\hdashline
SwinIR \cite{SwinIR}                   & $\times3$                      & 11.9M                       & 34.97                        & 0.9318                        & 30.93                        & 0.8534                        & 29.46                        & 0.8145                        & 29.75                        & 0.8826                        & 35.12                        & 0.9537                        \\
CAT-A \cite{CAT}                    & $\times3$                      & 16.6M                       & 35.06                        & 0.9326                        & 31.04                        & 0.8538                        & 29.52                        & 0.8160                        & 30.12                        & 0.8862                        & 35.38                        & 0.9546                        \\
DAT-S \cite{DAT}                    & $\times3$                      & 11.2M                       & 35.12                        & 0.9327                        & 31.04                        & 0.8543                        & 29.51                        & 0.8157                        & 29.98                        & 0.8846                        & 35.41                        & 0.9546                        \\
ART \cite{ART}                      & $\times3$                      & 16.6M                       & 35.07                        & 0.9325                        & 31.02                        & 0.8541                        & 29.51                        & 0.8159                        & 30.10                        & 0.8871                        & 35.39                        & 0.9548                        \\
HAT-S \cite{HAT}                    & $\times3$                      & 9.6M                       & 35.01                        & 0.9325                        & 31.05                        & 0.8550                        & 29.50                        & 0.8158                        & 30.15                        & 0.8879                        & 35.40                        & 0.9547                        \\
RGT-S \cite{RGT}                    & $\times3$                      & 10.2M                       & 35.11                        & 0.9328                        & 31.05                        & 0.8548                        & 29.53                        & 0.8164                        & 30.18                        & 0.8884                        & 35.39                        & 0.9548                        \\
\hdashline
MambaIR \cite{MambaIR}                  & $\times3$                      & 20.6M                       & 35.08                        & 0.9323                        & 30.99                        & 0.8536                        & 29.51                        & 0.8157                        & 29.93                        & 0.8841                        & 35.43                        & 0.9546                        \\
MambaIRv2-S \cite{MambaIRv2}              & $\times3$                      & 9.8M                        & 35.09                        & 0.9326                        & 31.07                        & 0.8547                        & 29.51                        & 0.8157                        & 30.08                        & 0.8871                        & 35.44                        & 0.9549                        \\
\hdashline
\rowcolor{gray!10} LSM-S (Ours)                   & $\times3$                      & 9.9M                        & {\color[HTML]{0000FF} 35.13}                        & {\color[HTML]{0000FF} 0.9329}                        & 31.03                        & 0.8545                        & 29.53                        & 0.8165                        & 30.20                        & 0.8889                        & 35.47                        & 0.9549                        \\
\rowcolor{gray!10} LSM (Ours)                   & $\times3$                      & 12.9M                        & \color[HTML]{0000FF}35.13                        & \color[HTML]{0000FF}0.9329                        & \color[HTML]{0000FF}31.11                        & \color[HTML]{0000FF}0.8552                        & \color[HTML]{0000FF}29.54                        & \color[HTML]{0000FF}0.8167                        & \color[HTML]{0000FF}30.28                        & \color[HTML]{0000FF}0.8901                        & \color[HTML]{0000FF}35.51                        & \color[HTML]{0000FF}0.9552                        \\
\rowcolor{gray!10} LSM+ (Ours)                   & $\times3$                     & 12.9M                        & \color[HTML]{FF0000}35.18                        & \color[HTML]{FF0000}0.9332                        & \color[HTML]{FF0000}31.18                        & \color[HTML]{FF0000}0.8560                        & \color[HTML]{FF0000}29.57                        & \color[HTML]{FF0000}0.8172                        & \color[HTML]{FF0000}30.42                        & \color[HTML]{FF0000}0.8916                        & \color[HTML]{FF0000}35.66                        & \color[HTML]{FF0000}0.9558                        \\
\hline
EDSR \cite{EDSR}                     & $\times4$                      & 43.0M                       & 32.46                        & 0.8968                        & 28.80                        & 0.7876                        & 27.71                        & 0.7420                        & 26.64                        & 0.8033                        & 31.02                        & 0.9148                        \\
\hdashline
SwinIR \cite{SwinIR}                   & $\times4$                      & 11.9M                       & 32.92                        & 0.9044                        & 29.09                        & 0.7950                        & 27.92                        & 0.7489                        & 27.45                        & 0.8254                        & 32.03                        & 0.9260                        \\
CAT-A \cite{CAT}                    & $\times4$                      & 16.6M                       & {\color[HTML]{FF0000} 33.08} & {\color[HTML]{0000FF} 0.9052} & 29.18                        & 0.7960                        & 27.99                        & 0.7510                        & 27.89                        & 0.8339                        & 32.39                        & 0.9285                        \\
DAT-S \cite{DAT}                    & $\times4$                      & 11.2M                       & 33.00                        & 0.9047                        & 29.20                        & 0.7962                        & 27.97                        & 0.7502                        & 27.68                        & 0.8300                        & 32.33                        & 0.9278                        \\
ART \cite{ART}                      & $\times4$                      & 16.6M                       & 33.04                        & 0.9051                        & 29.16                        & 0.7958                        & 27.97                        & 0.7510                        & 27.77                        & 0.8321                        & 32.31                        & 0.9283                        \\
HAT-S \cite{HAT}                    & $\times4$                      & 9.6M                       & 32.92                        & 0.9047                        & 29.15                        & 0.7958                        & 27.97                        & 0.7505                        & 27.87                        & 0.8346                        & 32.35                        & 0.9283                        \\
RGT-S \cite{RGT}                    & $\times4$                      & 10.2M                       & 32.98                        & 0.9047                        & 29.18                        & 0.7966                        & 27.98                        & 0.7509                        & 27.89                        & 0.8347                        & 32.38                        & 0.9281                        \\
\hdashline
MambaIR \cite{MambaIR}                  & $\times4$                      & 20.6M                       & 33.03                        & 0.9046                        & 29.20                        & 0.7961                        & 27.98                        & 0.7503                        & 27.68                        & 0.8287                        & 32.32                        & 0.9272                        \\
MambaIRv2-S \cite{MambaIRv2}              & $\times4$                      & 9.8M                        & 32.99                        & 0.9037                        & 29.23                        & 0.7965                        & 27.97                        & 0.7502                        & 27.73                        & 0.8307                        & 32.33                        & 0.9276                        \\
\hdashline
\rowcolor{gray!10} LSM-S (Ours)                   & $\times4$                      & 9.9M                        & 33.00                        & 0.9051                        & 29.20                        & 0.7963                        & 27.98                        & 0.7508                        & 27.88                        & 0.8348                        & 32.38                        & 0.9283                        \\
\rowcolor{gray!10} LSM (Ours)                   & $\times4$                      & 12.9M                       & 32.96                        & 0.9048                        & {\color[HTML]{0000FF} 29.24} & {\color[HTML]{0000FF} 0.7973} & {\color[HTML]{0000FF} 28.00} & {\color[HTML]{0000FF} 0.7511} & {\color[HTML]{0000FF} 27.94} & {\color[HTML]{0000FF} 0.8362} & {\color[HTML]{0000FF} 32.42} & {\color[HTML]{0000FF} 0.9285} \\
\rowcolor{gray!10} LSM+ (Ours)                  & $\times4$                      & 12.9M                       & {\color[HTML]{FF0000} 33.08} & {\color[HTML]{FF0000} 0.9053} & {\color[HTML]{FF0000} 29.30} & {\color[HTML]{FF0000} 0.7980} & {\color[HTML]{FF0000} 28.03} & {\color[HTML]{FF0000} 0.7517} & {\color[HTML]{FF0000} 28.07} & {\color[HTML]{FF0000} 0.8383} & {\color[HTML]{FF0000} 32.61} & {\color[HTML]{FF0000} 0.9297}\\
\hline
\end{tabular}%
}
\label{tab:qual_classic}
\vspace{5pt}
\end{table*}
\begin{table*}[]
\captionsetup{justification=centering}
\centering
\caption{Quantitative comparison on lightweight SR with state-of-the-art methods. FLOPs are measured at $1280 \times 720$ output resolution.}
\vspace{-5pt}
\resizebox{\textwidth}{!}{%
\footnotesize
\begin{tabular}{l|c|c|c|cc|cc|cc|cc|cc}
\hline
                         &                         &                           &                        & \multicolumn{2}{c|}{Set5}                                     & \multicolumn{2}{c|}{Set14}                                    & \multicolumn{2}{c|}{B100}                                     & \multicolumn{2}{c|}{Urban100}                                 & \multicolumn{2}{c}{Manga109}                                 \\
\multirow{-2}{*}{Method} & \multirow{-2}{*}{Scale} & \multirow{-2}{*}{\# Params} & \multirow{-2}{*}{FLOPs} & PSNR                         & SSIM                          & PSNR                         & SSIM                          & PSNR                         & SSIM                          & PSNR                         & SSIM                          & PSNR                         & SSIM                          \\
\hline
SwinIR-light \cite{SwinIR}             & $\times2$               & 910K                      & 244.2G                 & 38.14                        & 0.9611                        & 33.86                        & 0.9206                        & 32.31                        & 0.9012                        & 32.76                        & 0.9340                        & 39.12                        & 0.9783                        \\
ELAN-light \cite{ELAN}               & $\times2$               & 621K                      & 203.1G                 & 38.17                        & 0.9611                        & 33.94                        & 0.9207                        & 32.30                        & 0.9012                        & 32.76                        & 0.9340                        & 39.11                        & 0.9782                        \\
OmniSR \cite{OmniSR}                   & $\times2$               & 772K                      & 194.5G                 & 38.22                        & 0.9613                        & 33.98                        & 0.9210                        & {\color[HTML]{0000FF} 32.36} & {\color[HTML]{0000FF} 0.9020} & 33.05                        & 0.9363                        & 39.28                        & {\color[HTML]{0000FF} 0.9784} \\
\hdashline
MambaIR-light \cite{MambaIR}            & $\times2$               & 905K                      & 334.2G                 & 38.13                        & 0.9610                        & 33.95                        & 0.9208                        & 32.31                        & 0.9013                        & 32.85                        & 0.9349                        & 39.20                        & 0.9782                        \\
MambaIRv2-light \cite{MambaIRv2}          & $\times2$               & 774K                      & 286.3G                 & {\color[HTML]{0000FF} 38.26} & {\color[HTML]{FF0000} 0.9615} & {\color[HTML]{0000FF} 34.09} & {\color[HTML]{FF0000} 0.9221} & {\color[HTML]{0000FF} 32.36} & 0.9019                        & {\color[HTML]{FF0000} 33.26} & {\color[HTML]{0000FF} 0.9378} & {\color[HTML]{FF0000} 39.35} & {\color[HTML]{FF0000} 0.9785} \\
\hdashline
\rowcolor{gray!10} LSM-light (Ours)         & $\times2$               & 763K                      & 282.2G                 & {\color[HTML]{FF0000} 38.27} & {\color[HTML]{FF0000} 0.9615} & {\color[HTML]{FF0000} 34.14} & {\color[HTML]{0000FF} 0.9219} & {\color[HTML]{FF0000} 32.39} & {\color[HTML]{FF0000} 0.9023} & {\color[HTML]{0000FF} 33.24} & {\color[HTML]{FF0000} 0.9379} & {\color[HTML]{FF0000} 39.35} & {\color[HTML]{0000FF} 0.9784} \\
\hline
SwinIR-light \cite{SwinIR}             & $\times3$               & 918K                      & 110.8G                 & 34.62                        & 0.9289                        & 30.54                        & 0.8463                        & 29.20                        & 0.8082                        & 28.66                        & 0.8624                        & 33.98                        & 0.9478                        \\
ELAN-light \cite{ELAN}               & $\times3$               & 629K                      & 90.1G                  & 34.61                        & 0.9288                        & 30.55                        & 0.8463                        & 29.21                        & 0.8081                        & 28.69                        & 0.8624                        & 34.00                        & 0.9478                        \\
OmniSR \cite{OmniSR}                   & $\times3$               & 780K                      & 88.4G                  & 34.70                        & 0.9294                        & 30.57                        & 0.8469                        & \color[HTML]{0000FF} 29.28                        & 0.8094                        & 28.84                        & 0.8656                        & 34.22                        & 0.9487                        \\
\hdashline
MambaIR-light \cite{MambaIR}            & $\times3$               & 913K                      & 148.5G                 & 34.63                        & 0.9288                        & 30.54                        & 0.8459                        & 29.23                        & 0.8084                        & 28.70                        & 0.8631                        & 34.12                        & 0.9479                        \\
MambaIRv2-light \cite{MambaIRv2}          & $\times3$               & 781K                      & 126.7G                 & \color[HTML]{0000FF} 34.71                        & \color[HTML]{0000FF} 0.9298                        & \color[HTML]{FF0000} 30.68                        & \color[HTML]{0000FF} 0.8483                        & 29.26                        & \color[HTML]{0000FF} 0.8098                        & \color[HTML]{0000FF} 29.01                        & \color[HTML]{0000FF} 0.8689                        & \color[HTML]{FF0000} 34.41                        & \color[HTML]{0000FF} 0.9497                        \\
\hdashline
\rowcolor{gray!10} LSM-light (Ours)         & $\times3$               & 771K                      & 128.1G                 & \color[HTML]{FF0000} 34.76                        & \color[HTML]{FF0000} 0.9301                        & \color[HTML]{0000FF} 30.67                        & \color[HTML]{FF0000} 0.8485                        & \color[HTML]{FF0000} 29.30                        & \color[HTML]{FF0000} 0.8109                        & \color[HTML]{FF0000} 29.15                        & \color[HTML]{FF0000} 0.8712                        & \color[HTML]{FF0000} 34.41                        & \color[HTML]{FF0000} 0.9501                        \\
\hline
SwinIR-light \cite{SwinIR}             & $\times4$               & 930K                      & 63.6G                  & 32.44                        & 0.8976                        & 28.77                        & 0.7858                        & 27.69                        & 0.7406                        & 26.47                        & 0.7980                        & 30.92                        & 0.9151                        \\
ELAN-light \cite{ELAN}               & $\times4$               & 640K                      & 54.1G                  & 32.43                        & 0.8975                        & 28.78                        & 0.7858                        & 27.69                        & 0.7406                        & 26.54                        & 0.7982                        & 30.92                        & 0.9150                        \\
OmniSR \cite{OmniSR}                   & $\times4$               & 792K                      & 50.9G                  & 32.49                        & 0.8988                        & 28.78                        & 0.7859                        & 27.71                        & 0.7415                        & 26.64                        & 0.8018                        & 31.02                        & 0.9151                        \\
\hdashline
MambaIR-light \cite{MambaIR}            & $\times4$               & 924K                      & 84.6G                  & 32.42                        & 0.8977                        & 28.74                        & 0.7847                        & 27.68                        & 0.7400                        & 26.52                        & 0.7983                        & 30.94                        & 0.9135                        \\
MambaIRv2-light \cite{MambaIRv2}          & $\times4$               & 790K                      & 75.6G                  & \color[HTML]{0000FF} 32.51                        & \color[HTML]{0000FF} 0.8992                        & \color[HTML]{0000FF} 28.84                        & \color[HTML]{0000FF} 0.7878                        & \color[HTML]{0000FF} 27.75                        & \color[HTML]{0000FF} 0.7426                        & \color[HTML]{0000FF} 26.82                        & \color[HTML]{0000FF} 0.8079                        & \color[HTML]{0000FF} 31.24                        & \color[HTML]{0000FF} 0.9182                        \\
\hdashline
\rowcolor{gray!10} LSM-light (Ours)         & $\times4$               & 783K                      & 71.7G                  & \color[HTML]{FF0000} 32.55                      & \color[HTML]{FF0000} 0.8993                        & \color[HTML]{FF0000} 28.91                          & \color[HTML]{FF0000} 0.7886                        & \color[HTML]{FF0000} 27.77                        & \color[HTML]{FF0000} 0.7437                        & \color[HTML]{FF0000} 26.93                        & \color[HTML]{FF0000} 0.8106                        & \color[HTML]{FF0000} 31.35                        & \color[HTML]{FF0000} 0.9191                       \\
\hline
\end{tabular}%
}
\label{tab:qual_light}
\vspace{-5pt}
\end{table*}

\noindent \textbf{Initialization Parameters of LRU}
We investigate how eigenvalue initialization affects LRU performance by varying $r_\text{min}$, $r_\text{max}$, and $\theta_{max}$. The top row of \cref{fig:ablation} shows that eigenvalues form a ring-shaped distribution in the complex plane depending on these parameters, which determines the recurrence behavior of LRU. The first model sets $r_\text{min}$ and $r_\text{max}$ to 0.4 and 0.9, respectively, favoring local patterns. The second model reduces $\theta_{max}$ to $\pi/2$, encouraging low-frequency oscillations that favor more global receptive behavior. The last configuration represents our optimized setting. \cref{tab:initial} shows that lowering $r_\text{min}$ and $r_\text{max}$ leads to a performance drop of up to 0.19~dB due to reduced capacity for long-range modeling. Likewise, reducing the $\theta_{max}$ causes up to 0.12~dB degradation and visibly harms fine texture reconstruction as shown in the bottom row of \cref{fig:ablation}.
This trend differs from observations in prior LRU work, where smaller phase values were found to be beneficial. We conjecture that the difference is related to the behavior of SR, which must preserve both local structures and broader spatial consistency. These results suggest that eigenvalue initialization has a substantial impact in our architecture.
\setlength{\tabcolsep}{1pt}
\renewcommand{\arraystretch}{1}
\begin{figure*}[t]
    \footnotesize
    \captionsetup{justification=centering}
    \centering
    \resizebox{\textwidth}{!}{
    \begin{tabular}{ccccc p{0.8em} ccccc}
    
    \multirow{4}{*}{\parbox[t]{3cm}{\centering
        \includegraphics[width=3.01cm, valign=t]{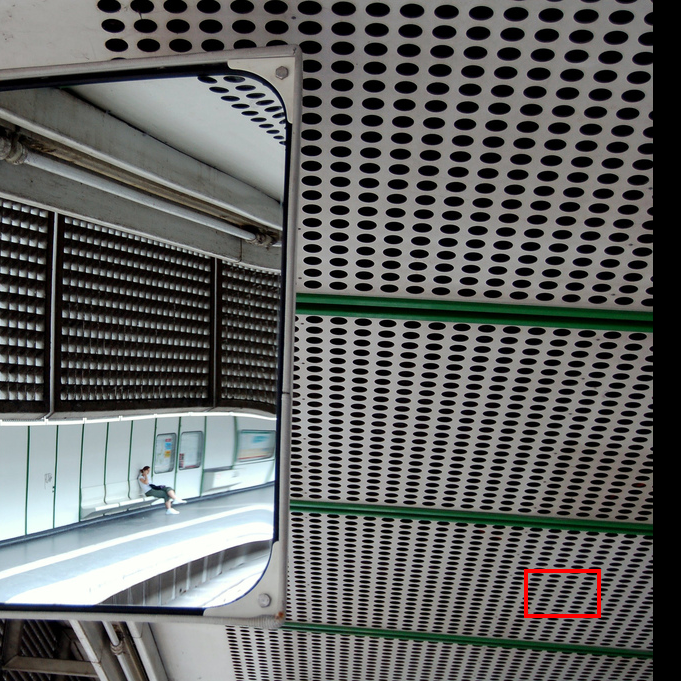}\\ 
        \vspace{0.2em}
        Urban100: img\_004
    }} & \includegraphics[width=2.12cm, valign=t]{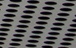} & \includegraphics[width=2.12cm, valign=t]{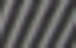} & \includegraphics[width=2.12cm, valign=t]{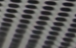} & \includegraphics[width=2.12cm, valign=t]{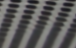} & &
    
    \multirow{4}{*}{\parbox[t]{3cm}{\centering
        \includegraphics[width=3.01cm, valign=t]{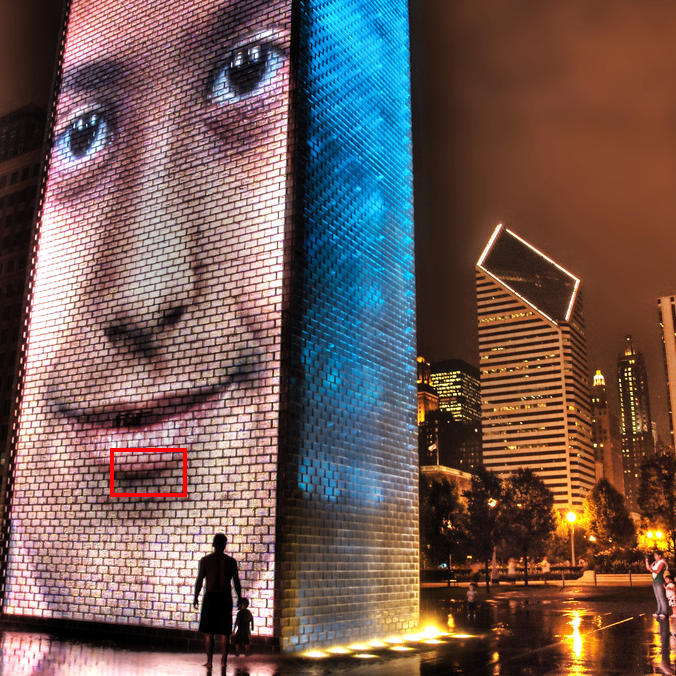}\\ 
        \vspace{0.2em}
        Urban100: img\_076
    }} & \includegraphics[width=2.12cm, valign=t]{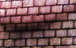} & \includegraphics[width=2.12cm, valign=t]{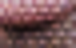} & \includegraphics[width=2.12cm, valign=t]{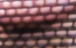} & \includegraphics[width=2.12cm, valign=t]{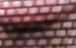} \\
    
    & HR & LR & SwinIR \cite{SwinIR} & MambaIR \cite{MambaIR} & &
    & HR & LR & SwinIR \cite{SwinIR} & MambaIR \cite{MambaIR} \\
    
    & \includegraphics[width=2.12cm, valign=t]{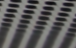} & \includegraphics[width=2.12cm, valign=t]{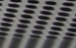} & \includegraphics[width=2.12cm, valign=t]{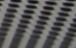} & \includegraphics[width=2.12cm, valign=t]{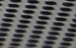} & &
    & \includegraphics[width=2.12cm, valign=t]{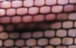} & \includegraphics[width=2.12cm, valign=t]{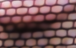} & \includegraphics[width=2.12cm, valign=t]{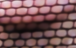} & \includegraphics[width=2.12cm, valign=t]{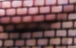} \\
    
    & CAT-A \cite{CAT} & ART \cite{ART} & MambaIRv2-S \cite{MambaIRv2} & LSM (Ours) & &
    & CAT-A \cite{CAT} & ART \cite{ART} & MambaIRv2-S \cite{MambaIRv2} & LSM (Ours) \\
    
    \rule{0pt}{0.5em} \\
    
    \multirow{4}{*}{\parbox[t]{3cm}{\centering
        \includegraphics[trim=10 10 10 10, clip, width=3.01cm, valign=t]{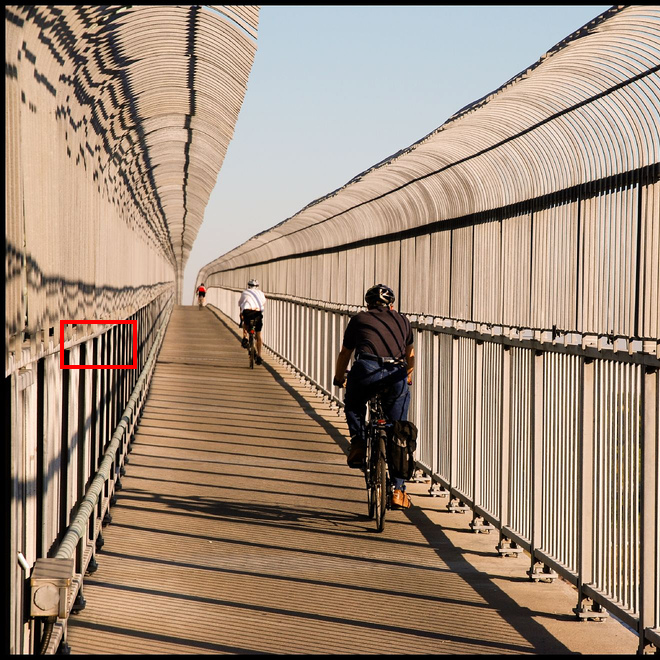}\\
        \vspace{0.2em}
        Urban100: img\_024
    }} & \includegraphics[width=2.12cm, valign=t]{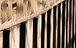} & \includegraphics[width=2.12cm, valign=t]{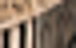} & \includegraphics[width=2.12cm, valign=t]{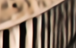} & \includegraphics[width=2.12cm, valign=t]{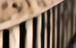} & &
    
    \multirow{4}{*}{\parbox[t]{3cm}{\centering
        \includegraphics[width=3.01cm, valign=t]{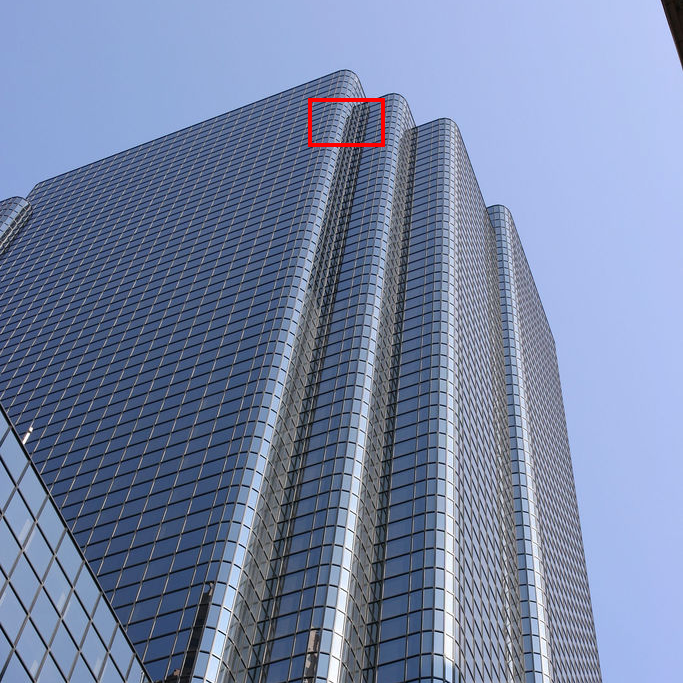}\\
        \vspace{0.2em}
        Urban100: img\_074
    }} & \includegraphics[width=2.12cm, valign=t]{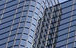} & \includegraphics[width=2.12cm, valign=t]{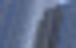} & \includegraphics[width=2.12cm, valign=t]{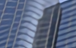} & \includegraphics[width=2.12cm, valign=t]{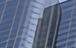} \\
        
    & HR & LR & SwinIR \cite{SwinIR} & MambaIR \cite{MambaIR} & &
    & HR & LR & SwinIR \cite{SwinIR} & MambaIR \cite{MambaIR} \\
        
    & \includegraphics[width=2.12cm, valign=t]{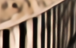} & \includegraphics[width=2.12cm, valign=t]{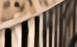} & \includegraphics[width=2.12cm, valign=t]{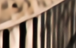} & \includegraphics[width=2.12cm, valign=t]{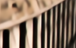} & &
    & \includegraphics[width=2.12cm, valign=t]{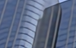} & \includegraphics[width=2.12cm, valign=t]{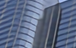} & \includegraphics[width=2.12cm, valign=t]{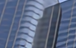} & \includegraphics[width=2.12cm, valign=t]{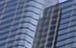} \\
        
    & CAT-A \cite{CAT} & ART \cite{ART} & MambaIRv2-S \cite{MambaIRv2} & LSM (Ours) & &
    & CAT-A \cite{CAT} & ART \cite{ART} & MambaIRv2-S \cite{MambaIRv2} & LSM (Ours) \\
    
    \end{tabular}%
    }
    \vspace{-5pt}
    \caption{Qualitative comparisons of LSM with other challenging methods on $\times 4$ classic SR.}
    \label{fig:qual}
\vspace*{-18pt}
\end{figure*}

\noindent \textbf{Multi-role of SMU}
We evaluate four variants to analyze the multi-role design of the SMU. The baseline completely removes all SMU roles and uses only the WMS Block with a vanilla LRU. The second variant adds the semantic categorization via the external dictionary, yielding slight performance gains on Manga109 as shown in \cref{tab:multirole}. The third includes the cross-attention layer as a secondary functional role, and the final version enables LRU modulation as the main role. The full model with modulating tokens $\mathbf{M}_k$ improves PSNR by 0.12~dB on Set14 and 0.08~dB on Manga109 over the baseline, demonstrating the synergy of the multi-role design.

\noindent \textbf{Effect of Modulating Tokens}
We further analyze the impact of modulating tokens $\mathbf{M}_k$. In \cref{tab:M_t}, the first model disables all modulation paths, keeping only cross-attention. In the next three rows, each switch for $\lambda$, $B$, and $C$ is individually enabled. All cases show improvements on at least one dataset, confirming the utility of each modulation path. Since modulating tokens share a softmax branch with cross-attention, disabling a switch does not nullify that channel entirely but biases training toward the cross-attention role. Consequently, the performance is slightly reduced when modulation is removed in some datasets. In the last row, we decouple modulation from the softmax by introducing a separate linear layer. This results in up to 0.10~dB PSNR drop, indicating that the coupled design promotes an effective balance between cross-attention and modulation while maintaining parameter efficiency.

\subsection{Comparisons with State-of-the-Art Methods}
For testing, we adopt five standard benchmark datasets: Set5 \cite{Set5}, Set14 \cite{Set14}, B100 \cite{B100}, Urban100 \cite{Urban100}, and Manga109 \cite{Manga109}. We evaluate performance using PSNR and SSIM \cite{SSIM}, consistent with prior works.

\noindent \textbf{Quantitative Results} We first compare our models with state-of-the-art classic SR methods. In \cref{tab:qual_classic}, we group models by backbone type using dashed lines, with EDSR \cite{EDSR} as a CNN-based model, SwinIR \cite{SwinIR}, CAT \cite{CAT}, DAT \cite{DAT}, ART \cite{ART}, HAT \cite{HAT}, and RGT \cite{RGT} as Transformer-based models, MambaIR \cite{MambaIR} and MambaIRv2 \cite{MambaIRv2} as Mamba-based models, and our LSM-S and LSM as LRU-based models. Following prior works \cite{EDSR, RCAN}, we adopt a self-ensemble strategy during testing and denote ensembled models with a ``+'' suffix. We focus on recent efficient Transformer-based models under 20M parameters, commonly categorized as small or medium-sized. 
Despite having fewer parameters, our LSM models achieve the best performance across all scales on five benchmark datasets. Compared to Mamba-based methods, LSM-S surpasses MambaIRv2-S by 0.15~dB on $\times4$ Urban100 in \cref{tab:qual_classic} where modeling long-range dependencies is essential. This demonstrates the effectiveness of our design, which integrates an efficient LRU backbone with initialization of eigenvalues (\cref{fig:ablation}, \cref{tab:initial}) and modulating tokens (\cref{tab:M_t}) to better capture high-frequency details and repetitive patterns.
For the lightweight SR task, we compare LSM-light with SwinIR-light \cite{SwinIR}, ELAN-light \cite{ELAN}, and OmniSR \cite{OmniSR} as Transformer-based models, and MambaIR-light \cite{MambaIR} and MambaIRv2-light \cite{MambaIRv2} as Mamba-based models. As shown in \cref{tab:qual_light}, LSM-light outperforms recent models of similar model size, improving over OmniSR by 0.32~dB and MambaIRv2-light by 0.10~dB on $\times4$ Manga109. Notably, the performance margin becomes larger on Urban100 and Manga109 at higher scales, where learning global information plays a more critical role than in $\times2$ SR.\\
\begin{table}[!t]
\vspace{3pt}
\captionsetup{justification=centering}
\centering
\caption{Model size and computational complexity comparison.}
\begin{subtable}[t]{0.48\textwidth}
\vspace{-8pt}
\centering
\caption{Comparison with Mamba methods on $\times4$ classic SR.}
\footnotesize
\vspace{-1pt}
\begin{tblr}{
  width=\linewidth,
  colspec={X[l]|c|c|cc|cc},
  colsep=2.5pt, rowsep=1pt,
  row{1-2} = {font=\footnotesize}, rows = {font=\footnotesize}
}
\toprule 
\SetCell[r=2]{c} Method & \SetCell[r=2]{c} \# Params & \SetCell[r=2]{c} FLOPs & \SetCell[c=2]{c} Urban100 & & \SetCell[c=2]{c} Manga109 & \\
 &  &  & PSNR & SSIM & PSNR & SSIM \\
\midrule
MambaIR \cite{MambaIR}     & 20.6M & 394.6G & 27.68 & 0.8287 & 32.32 & 0.9272 \\
MambaIRv2-S \cite{MambaIRv2}  & \textbf{9.8M} & \textbf{202.9G} & 27.73 & 0.8307 & 32.33 & 0.9276 \\
LSM-S (Ours) & \textbf{9.9M} & \textbf{203.5G} & 27.88 & 0.8348 & 32.38 & 0.9283 \\
LSM (Ours)   & 12.9M & 265.0G & \textbf{27.94} & \textbf{0.8362} & \textbf{32.42} & \textbf{0.9285} \\
\bottomrule
\end{tblr}
\label{tab:computation_a}
\end{subtable}
\hfill
\begin{subtable}[t]{0.48\textwidth}
\centering
\footnotesize
\vspace{3pt}
\caption{Comparison with efficient Transformer methods on $\times2$ classic SR.}
\begin{tblr}{
  width=\linewidth,
  colspec={X[l]|c|c|cc|cc},
  colsep=3pt, rowsep=1pt,
  row{1-2} = {font=\footnotesize}, rows = {font=\footnotesize}
}
\toprule
\SetCell[r=2]{c} Method & \SetCell[r=2]{c} \# Params & \SetCell[r=2]{c} FLOPs & \SetCell[c=2]{c} Urban100 & & \SetCell[c=2]{c} Manga109 & \\
 &  &  & PSNR & SSIM & PSNR & SSIM \\
\midrule
SwinIR \cite{SwinIR}       & 11.8M & 205.3G        & 33.81 & 0.9427 & 39.92 & 0.9797 \\
CAT-A \cite{CAT}       & 16.5M & 350.7G        & 34.26 & 0.9440 & 40.10 & 0.9805 \\
RGT-S \cite{RGT}       & 10.1M & \textbf{183.1G} & 34.32 & 0.9457 & 40.18 & 0.9805 \\
LSM-S (Ours) & \textbf{9.7M} & 193.5G & 34.40 & 0.9464 & 40.25 & 0.9805 \\
LSM (Ours)   & 12.8M & 255.0G       & \textbf{34.43} & \textbf{0.9466} & \textbf{40.35} & \textbf{0.9809} \\
\bottomrule 
\end{tblr}
\label{tab:computation_b}
\end{subtable}
\label{tab:computation}
\vspace{-18pt}
\end{table}
\noindent \textbf{Qualitative Results} \cref{fig:qual} shows visual comparisons with representative methods on the $\times4$ Classic SR task. Most competing methods struggle to recover sharp textures in challenging regions. For example, in \textit{img\_004} and \textit{img\_024}, many methods fail to reconstruct circular or striped patterns. In contrast, our method restores these textures more accurately and with fewer artifacts. This is attributed to the category-based modulated LRU (CML), which enables the model to capture similar textures across the entire image and thereby enhance global consistency.
\subsection{Comparisons of Model Size and Complexity}
We compare our model with recent SR methods in terms of SR performance (PSNR, SSIM \cite{SSIM} on Urban100 \cite{Urban100} and Manga109 \cite{Manga109}), model size (\# Params), and computational cost (FLOPs) evaluated with an input size of $3 \times 128 \times 128$ from two main perspectives.\\
\noindent \textbf{Comparison with Mamba Backbones}
We compare our models with Mamba-based methods including MambaIR \cite{MambaIR} and the latest MambaIRv2 \cite{MambaIRv2} to evaluate the trade-off between long-sequence modeling and efficiency of the LRU backbone. As shown in \cref{tab:computation_a}, LSM reduces FLOPs by 32.8\% and improves PSNR by 0.26~dB over MambaIR on Urban100. These results highlight the efficiency and accuracy achieved by our model within the SSM variant.\\
\noindent \textbf{Comparison with Transformer Backbones}
We compare our models with efficient Transformer-based methods including SwinIR \cite{SwinIR}, CAT \cite{CAT}, and RGT \cite{RGT} which are proposed to alleviate the quadratic complexity of standard self-attention.
As shown in \cref{tab:computation_b}, LSM reduces parameters by 41.2\% and FLOPs by 44.8\% relative to CAT-A while improving PSNR by 0.14~dB and 0.15~dB on Urban100 and Manga109, respectively. These results demonstrate the competitiveness of our LRU-based model against linear-complexity Transformer baselines.

\section{Discussion}
\label{sec:discussion}
\begin{figure}[t]
\scriptsize
\centering
\hspace{-18pt}

\hspace{0pt}
\stackunder[2pt]{\includegraphics[trim=0 0 0 55,clip,width = 1.05in]{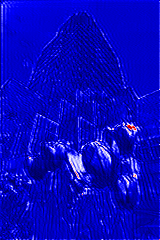}}{Vanilla LRU}
\hspace{0pt}%
\stackunder[2pt]{\includegraphics[trim=0 0 0 55,clip,width = 1.05in]{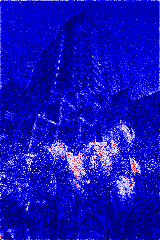}}{+ Categorize}
\hspace{0pt}%
\stackunder[2pt]{\includegraphics[trim=0 0 0 55,clip,width = 1.05in]{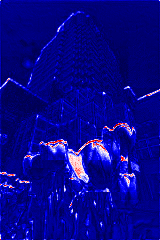}}{+ Categorize \& Modulate}

\vspace*{-7pt}
\caption{Visualization of hidden states. Semantic modulation enhances global structure and local detail over vanilla LRU.}
\label{fig:hidden_state}
\vspace*{-15pt}
\end{figure}

\begin{figure}[t]
    \centering
    \includegraphics[trim=11 11 9 11 ,clip,width = 3.1in]{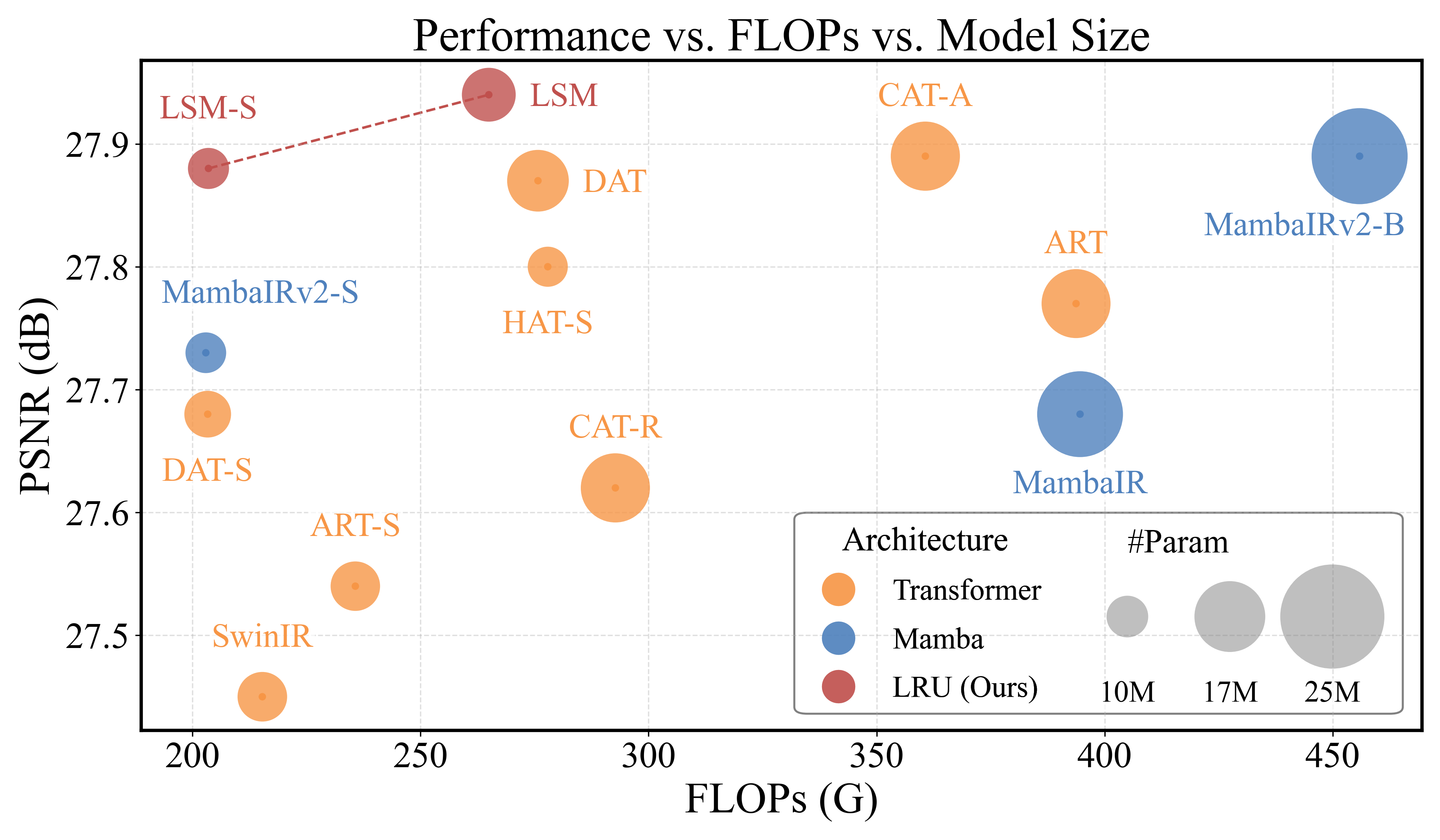}
    \vspace{-7pt}
    \caption{Comparison of PSNR and FLOPs on $\times4$ classic SR using the Urban100 dataset \cite{Urban100}.}
    \label{fig:computation}
    \vspace{-5pt}
\end{figure}

\noindent \textbf{Modulation Effects on Hidden States}
In \cref{fig:hidden_state}, we compare hidden states from three variants to demonstrate the effectiveness of semantic modulation in the proposed LSM: vanilla LRU, categorized LRU, and modulated LRU with categorization. Since channel orders differ across models, we select the most similar channels based on cosine similarity and visualize them after normalization. The vanilla LRU fails to capture key textures such as building facades and flower stems. Categorization partially resolves this, but fine details remain blurry. The final modulated model improves both global structure and local detail.\\
\noindent \textbf{Complexity and Performance}
We present an additional comparison of computational complexity and performance for representative SR models based on Transformer and Mamba architectures as shown in \cref{fig:computation}. Our methods which employ an LRU backbone achieve a better trade-off between computational complexity and SR performance. Notably, our LSM model achieves 41.8\% lower FLOPs and 43.9\% fewer parameters compared to MambaIRv2-B, while still achieving higher PSNR performance on $\times4$ scale with Urban100 dataset.\\
\noindent \textbf{Turning Memory Efficiency into Performance}
In \cref{fig:memory}, our model is significantly more memory-efficient by utilizing the LRU backbone compared to Transformer and Mamba backbones with linear complexity. This allows LSM to handle larger input resolutions under limited hardware resources. In the dictionary fine-tuning stage, we leverage this property to maximize utilization of this margin within a given memory budget, achieving strong performance with relatively low computational complexity.
\begin{figure}[t]
    \captionsetup{justification=justified,singlelinecheck=false}
    \centering
    \includegraphics[trim=7 11 7 0 ,clip,width = 3.0in]{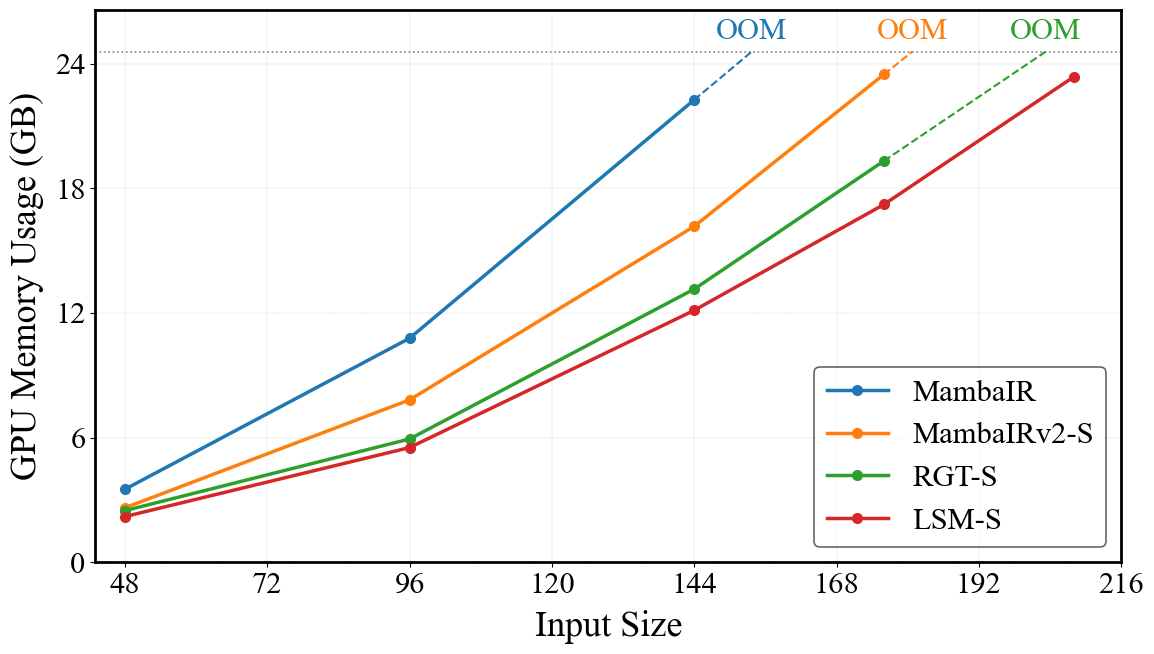}

    \vspace{-7pt}
    \caption{GPU memory usage per input resolution during training.}
    \label{fig:memory}
   \vspace{-15pt}
\end{figure}
\section{Conclusion}
We proposed LSM, a novel LRU-based SR network that preserves the stability of LRU while enhancing reconstruction quality. To the best of our knowledge, we are the first to utilize LRUs for SR. By applying pixel-wise modulation to the transition matrix via semantic modulating unit (SMU), LSM successfully captures both long-range contextual information and detailed local features. 
Furthermore, the proposed SMU improves the static nature of the scanning core via dictionary learning, which facilitates pre-categorization and enhances feature representation. Extensive experiments demonstrate that our network outperforms existing models in both efficiency and performance, showing strong potential as a next-generation SR backbone. 

\noindent \textbf{Acknowledgements} 
This work was supported by the National Research Foundation of Korea (NRF) grant funded by the Korea government
(MSIT) (RS-2024-00335741).
{
    \small
    \bibliographystyle{ieeenat_fullname}
    \bibliography{main}
}

\maketitlesupplementary

\section*{A. Training Settings}
\noindent \textbf{Classic SR}
Following previous works \cite{SwinIR, EDSR}, we use DIV2K \cite{DIV2K} and Flickr2K \cite{EDSR} as the training datasets. 
We train with batch size 32. Patches are augmented by random flips and $90^\circ$, $180^\circ$, $270^\circ$ rotations. Training proceeds in two steps. In the first step, inputs are cropped to $64\times 64$, and we minimize the $\ell_1$ pixel loss using AdamW \cite{AdamW} with $\beta_1=0.9$, $\beta_2=0.9$. For $\times 2$ upscaling, training runs for $300\text{k}$ iterations with initial learning rate $2\times 10^{-4}$, halved at the $250\text{k}$ milestone. In the subsequent fine-tuning step, following previous work \cite{ATD}, we use larger patches ($96\times 96$ for LSM‑S, $92\times 92$ for LSM) chosen for NVIDIA RTX 3090 GPU capacity to better exploit the semantic modulating unit (SMU) and memory efficiency. The training runs for $200\text{k}$ iterations and the same initial learning rate is used with halving at milestones. Total training is $500\text{k}$ iterations. For $\times 3$ and $\times 4$, we skip first step for efficiency, initialize from $\times 2$ weights, and apply only fine-tuning step for 250k iterations. A $10\text{k}$ warm-up at each step increases the learning rate linearly from $0$ to the initial value. \\
\noindent \textbf{Lightweight SR}
In the LSM-light model, only the DIV2K \cite{DIV2K} dataset is used for training unlike the classic SR. To match the batch size with previous works \cite{ELAN, OmniSR, MambaIRv2}, we doubled it compared to the classic SR setting, while keeping all other training strategies identical to those of LSM-S.

\section*{B. Additional Quantitative Comparison}
Our objective is to propose an efficient SR backbone based on LRU, a lightweight SSM variant.
To this end, all models were trained under practical compute constraints, using 24GB of GPU memory across 8 GPUs.
Accordingly, the main paper primarily compares our model with existing small size baselines that adopt Transformer and Mamba backbones. We further demonstrate the potential of our model as a new SR backbone by evaluating it on larger models and higher-resolution datasets in terms of performance and efficiency.

\noindent \textbf{Comparison with Large Models}
We compare our model with two recent larger models on $\times 4$ SR: Transformer-based HAT \cite{HAT} and Mamba-based MambaIRv2-B \cite{MambaIRv2}. As shown in \cref{tab:qual_add1}, our model achieves competitive performance despite reducing the number of parameters and FLOPs significantly by 38\% and 36\% compared to HAT, and by 44\% and 42\% compared to MambaIRv2-B, respectively.

\noindent \textbf{Comparison with Dictionary-based Model}
ATD \cite{ATD}, which inspired our approach, employs category-based attention with a parallel architecture. While minimizing parameter overhead, it achieves strong performance gains. The dictionary operation used in ATD plays a key role in overcoming the spatial limitations of local attention. Similarly, we reinterpret the mechanism by mitigating the single-scan limitation of LRU in our model. Futhermore we assign it the role of computationally efficient modulation. We compare the ATD-light and LSM-light models at the $\times 2$ scale in \cref{tab:qual_add2}. Despite a similar number of parameters due to the parallel structure of ATD, our model reduces FLOPs by 35\% and achieves 1.5$\times$ lower latency, while maintaining comparable performance.
This result supports the validity of our approach, where the integration of dynamic modulation enhances suitability for SR tasks by extending long-range modeling capacity of LRU.

\noindent \textbf{Comparison on high-resolution datasets}
Our work emphasizes that the carefully designed initialization of the LRU provides a foundational basis for its recurrence behavior, which is essential for effective long-range modeling.  
To further validate this claim, we conduct additional experiments on high-resolution datasets \cite{Test8K}.  
As shown in \cref{tab:qual_add3}, our LSM-S consistently outperforms SwinIR \cite{SwinIR} and MambaIRv2-S \cite{MambaIRv2} across most datasets despite its efficient computational complexity.
\begin{table}[]
\centering
\vspace{3pt}
\renewcommand{\thetable}{B.1}
\setlength{\tabcolsep}{1.2pt}
\captionsetup{justification=centering}
\caption{Quantitative comparison with large size models}
\vspace{-8pt}
\label{tab:qual_add1}
\scriptsize
\resizebox{\columnwidth}{!}{%
\begin{tabular}{l|c|c|c|c|c|c|c}
\hline
Method      & \# Params       & FLOPs        & Set5           & Set14          & B100           & Ub.100       & Mg.109       \\
\hline
HAT         & 20.8M          & 412G          & 33.04          & 29.23          & \textbf{28.00} & \textbf{27.97} & 32.48          \\
MambaIRv2-B & 23.1M          & 455G          & \textbf{33.14} & 29.23          & \textbf{28.00} & 27.89          & \textbf{32.57} \\
LSM  & \textbf{12.9M} & \textbf{265G} & 32.96          & \textbf{29.24} & \textbf{28.00} & 27.94          & 32.42         \\
\hline
\end{tabular}%
}
\vspace{-10pt}
\end{table}
\begin{table}[]
\centering
\vspace{0pt}
\renewcommand{\thetable}{B.2}
\setlength{\tabcolsep}{1.2pt}
\captionsetup{justification=centering}
\caption{Quantitative comparison with dictionary-based model}
\vspace{-8pt}
\label{tab:qual_add2}
\scriptsize
\resizebox{\columnwidth}{!}{%
\begin{tabular}{l|c|c|c|c|c|c|c|c|c}
\hline
Method                     & Latency & \# Params                        & FLOPs                 & Metric & Set5           & Set14          & B100           & Ub.100       & Mg.109       \\
\hline
\multirow{2}{*}{ATD-light} & \multirow{2}{*}{914ms} & \multirow{2}{*}{\textbf{753K}} & \multirow{2}{*}{380G} &
PSNR  & \textbf{38.29} & 34.10          & \textbf{32.39} & \textbf{33.27} & \textbf{39.52} \\
                           &                                &                       &                       &
SSIM & \textbf{0.9616} & 0.9217         & \textbf{0.9023} & 0.9375         & \textbf{0.9789} \\
\multirow{2}{*}{LSM-light} & \multirow{2}{*}{\textbf{611ms}} 
  & \multirow{2}{*}{763K}
  & \multirow{2}{*}{\textbf{282G}}
  & PSNR & 38.27 & \textbf{34.14} & \textbf{32.39} & 33.24         & 39.35 \\

  &       &                                    &                      & SSIM & 0.9615 & \textbf{0.9219} & \textbf{0.9023} & \textbf{0.9379} & 0.9784 \\
\hline
\end{tabular}%
}
\vspace{-10pt}
\end{table}
\begin{table}[]
\centering
\vspace{0pt}
\renewcommand{\thetable}{B.3}
\setlength{\tabcolsep}{1.2pt}
\captionsetup{justification=centering}
\caption{Quantitative comparison on high-resolution datasets}
\vspace{-8pt}
\label{tab:qual_add3}
\scriptsize
\resizebox{\columnwidth}{!}{%
\begin{tabular}{l|c|c|cc|cc|cc}
\hline
\multirow{2}{*}{Method} & \multirow{2}{*}{\# Params} & \multirow{2}{*}{FLOPs} & \multicolumn{2}{c|}{Test2k} & \multicolumn{2}{c|}{Test4k} & \multicolumn{2}{c}{Test8k} \\
                               &                                &                                 & PSNR    & SSIM   & PSNR     & SSIM     & PSNR     & SSIM     \\
\hline
SwinIR                            & 11.9M                            & 215.3G                              & 27.99            & 0.7898          & 29.48             & 0.8349            & 35.57             & 0.9034            \\
MambaIRv2-S                            & \textbf{9.8M}                           & \textbf{202.9G}                             & 28.07            & 0.7909           & 29.56             & 0.8359            & \textbf{35.74}             & 0.9047            \\
LSM-S                            & \textbf{9.9M}                           & \textbf{203.5G}                              & \textbf{28.11}            & \textbf{0.7924}          & \textbf{29.60}             & \textbf{0.8371}            & 35.73             & \textbf{0.9050}           \\
\hline
\end{tabular}%
}
\vspace{-15pt}
\end{table}

\section*{C. Preliminaries}
LRUs \cite{LRU} serve as the core backbone of this study. Unlike conventional RNNs that struggle with long-sequence learning due to vanishing and exploding gradient problems and the inefficiency of sequential computation, LRUs demonstrate strong performance in long-range dependency modeling and high computational efficiency through a series of structural changes and initialization strategies. We provide a full derivation of the LRU formulation presented in the methodology section of the main paper. \\
\noindent \textbf{Vanilla RNN}
A standard RNN layer \cite{RNN1} consumes an $H_{\text{in}}$-dimensional input, produces an $N$-dimensional hidden state and an $H_{\text{out}}$-dimensional output, and typically includes a non-linear activation function $\sigma$:
\begin{equation}
\begin{split}
h_k &= \sigma(A h_{k-1} + B u_k), \\
y_k &= C h_k + D u_k,
\end{split}\tag{C.1}
\end{equation}
where $A \in \mathbb{R}^{N \times N}$, $B \in \mathbb{R}^{N \times H_{\text{in}}}$, $C \in \mathbb{R}^{H_{\text{out}} \times N}$, and $D \in \mathbb{R}^{H_{\text{out}} \times H_{\text{in}}}$ are trainable, and $h_0 = 0$.\\
\noindent \textbf{Linearizing Recurrences}
The first key modification in LRU is the removal of the non-linearity $\sigma$ from the hidden state update, opting for a linear recurrence. This enhances learning stability and enables parallelization without sacrificing model expressivity. The overall non-linearity is instead provided by Multi-Layer Perceptron (MLP) or Gated Linear Unit (GLU) blocks placed between each LRU block:
\begin{equation}
\begin{split}
h_k &= A h_{k-1} + B u_k, \\
y_k &= C h_k + D u_k,
\end{split}\tag{C.2}
\end{equation}
which unrolls as $h_k = A^k h_0 + \sum_{j=0}^{k-1} A^j B\, u_{k-j}$.  In long sequences, the hidden state can explode or vanish depending on the magnitude of the eigenvalues of matrix $A$.

\noindent \textbf{Complex Diagonal Recurrences}
To maximize the computational efficiency of the linear recurrence, matrix $A$ is reparameterized as a complex-valued diagonal matrix $\Lambda$. This leverages the eigendecomposition of $A$, expressed as $A=P\Lambda P^{-1}$. In the eigen-basis $\bar{h}_k=P^{-1}h_k$, the hidden state can be linearly expressed as:
\begin{equation}
\begin{split}
\bar{h}_k &= \Lambda \bar{h}_{k-1} + \bar{B}\, u_k, \\
\bar{y}_k &= \bar{C}\, \bar{h}_k + D u_k,
\end{split}\tag{C.3}
\end{equation}
where $\bar{B} = P^{-1} B$ and $\bar{C} = C P$. Then $\bar{h}_k = \Lambda^k \bar{h}_0 + \sum_{m=0}^{k-1} \Lambda^m \bar{B}\, u_{k-m}$ with elementwise powers on the diagonal of $\Lambda$, which is parallel-scan friendly.

\noindent \textbf{Stable Exponential Parameterization}
LRU enhances learning stability and strengthens long-range dependency modeling by controlling the eigenvalue $\lambda$ distribution of the recurrent matrix, rather than relying on a specific deterministic initialization. Exponential parameterization is used to control the magnitude and phase of eigenvalues, which effectively separates them to improve the performance of optimizers:
\begin{equation}
\begin{split}
\Lambda &= \operatorname{diag}(\lambda), \\
\lambda_j &= \exp\!\big(-\exp(\nu^{\log}_j)\big)\; \exp\!\big(i\, \exp(\theta^{\log}_j)\big),
\end{split}\tag{C.4}
\end{equation}
where j refers to the index of each individual eigenvalue $\lambda_j$, which comes with trainable $\nu^{\log}_j$, $\theta^{\log}_j \in \mathbb{R}$. For initialization, $\lambda_j$ are sampled to be uniformly distributed on an annulus in the complex plane, defined by inner radius $r_{\min}$ and outer radius $r_{\max}$. The phase of $\lambda_j$ is uniformly sampled within a specified range, typically [0, 2$\pi$] or a smaller slice for tasks requiring very long-range reasoning. Specifically, the trainable parameters $\nu^{\log}_j$ and $\theta^{\log}_j$ are initialized using independent uniform random variables $u_1$, $u_2$ $\in$ [0, 1] as follows:
\begin{equation}
\begin{split}
\nu^{\log}_j &= \log\left(-\frac{1}{2} \log\left(u_1(r_{\max}^2 - r_{\min}^2) + r_{\min}^2\right)\right),\\
\theta^{\log}_j &= \log(\theta_{max} u_2),
\end{split}\tag{C.5}
\end{equation}
where $\theta_{max}$ defines the upper limit of the phase sampling range. This initialization strategy sets an effective dependency range for each $\lambda_j$ and the results are further analyzed in the ablation section of the main paper.    

\noindent \textbf{Normalization}
To prevent hidden activation blow-up when $|\lambda_j|$ is close to one, LRU introduces a forward normalization factor $\gamma_j=\sqrt{1-|\lambda_j|^2}$ applied channelwise. The modified hidden state update and output equations are as follows:
\begin{equation}
\begin{split}
\bar{h}_k &= \operatorname{diag}(\lambda) \odot \bar{h}_{k-1} + \gamma \odot (\bar{B}\, u_k), \\
\bar{y}_k &= \bar{C}\, \bar{h}_k + D u_k,
\end{split}\tag{C.6}
\end{equation}
where $\gamma=\operatorname{diag}(\gamma_j)$ broadcasts across channels and $\odot$ denotes elementwise multiplication.

\section*{D. Additional Visual Results}
To further support the findings presented in the main paper, we provide additional qualitative visualizations.

\noindent \textbf{Visualization of hidden states}
We further visualize the modulation effects on hidden states in \cref{fig:add_visual} to demonstrate the consistency of our findings.
Following the same strategy, we sort channels across different models based on cosine similarity to highlight similarly activated responses.  
The results show that the vanilla LRU struggles to capture key textures such as the bird's beak and window patterns.  
With semantic categorization, hidden states exhibit more coherent activation across spatially distant pixels with similar meanings.  
Finally, applying the modulated LRU to categorized pixels allows the model to balance long-range semantic consistency and local texture, yielding the most faithful representations among variants.

\noindent \textbf{Visualization of categorization} 
We visualize the categorization results before feeding into the LRU in \cref{fig:add_category}.

\begin{figure*}[t]
    \scriptsize
    \renewcommand{\thefigure}{D.3}
    \captionsetup{justification=centering}
    \centering
    \resizebox{\textwidth}{!}{
    \begin{tabular}{ccccc}
    
    \multirow{4}{*}{\parbox[t]{3cm}{\centering
        \includegraphics[width=2.96cm, valign=t]{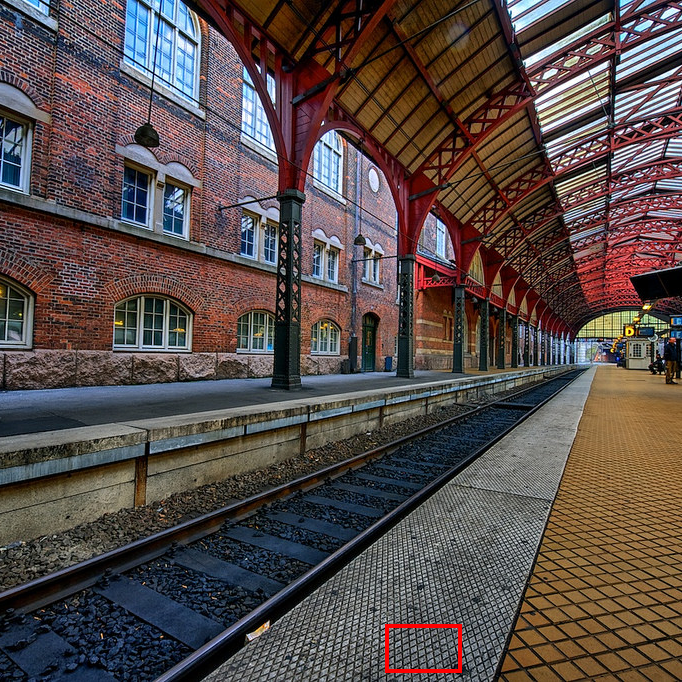}\\
        Urban100: img\_098
    }} & \includegraphics[width=2.12cm, valign=t]{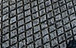} & \includegraphics[width=2.12cm, valign=t]{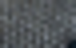} & \includegraphics[width=2.12cm, valign=t]{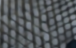} & \includegraphics[width=2.12cm, valign=t]{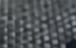} \\
    & HR & LR & SwinIR & MambaIR \\
    & \includegraphics[width=2.12cm, valign=t]{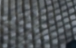} & \includegraphics[width=2.12cm, valign=t]{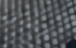} & \includegraphics[width=2.12cm, valign=t]{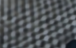} & \includegraphics[width=2.12cm, valign=t]{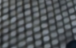} \\
    & CAT-A & ART & MambaIRv2-S & LSM (Ours) \\
    \rule{0pt}{1.5em}
    
    \multirow{4}{*}{\parbox[t]{3cm}{\centering
        \includegraphics[width=2.96cm, valign=t]{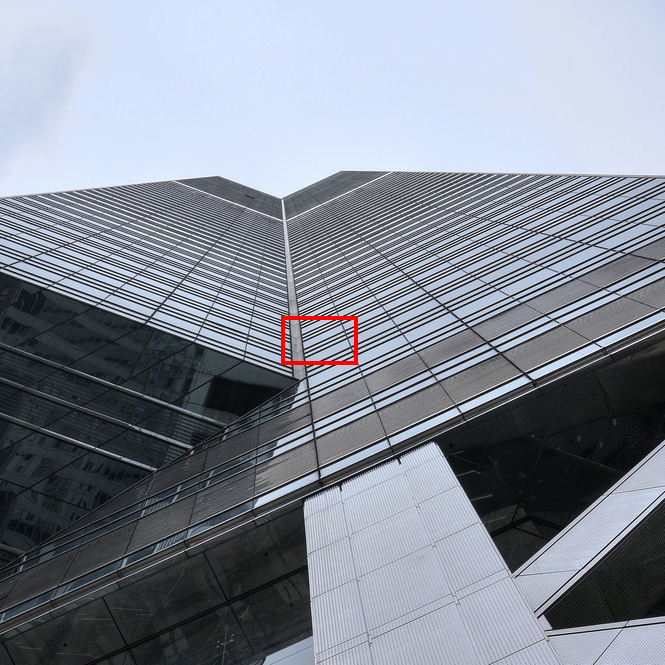}\\
        Urban100: img\_059
    }} & \includegraphics[width=2.12cm, valign=t]{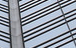} & \includegraphics[width=2.12cm, valign=t]{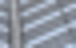} & \includegraphics[width=2.12cm, valign=t]{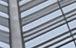} & \includegraphics[width=2.12cm, valign=t]{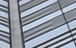} \\
    & HR & LR & SwinIR & MambaIR \\
    & \includegraphics[width=2.12cm, valign=t]{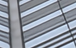} & \includegraphics[width=2.12cm, valign=t]{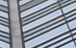} & \includegraphics[width=2.12cm, valign=t]{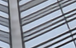} & \includegraphics[width=2.12cm, valign=t]{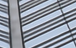} \\
    & CAT-A & ART & MambaIRv2-S & LSM (Ours) \\
    \rule{0pt}{1.5em}
    
    \multirow{4}{*}{\parbox[t]{3cm}{\centering
        \includegraphics[width=2.96cm, valign=t]{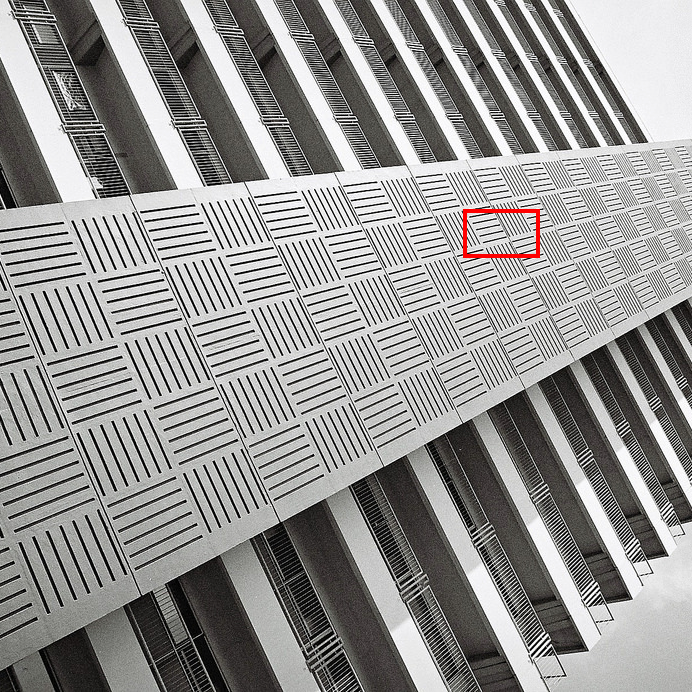}\\
        Urban100: img\_092
    }} & \includegraphics[width=2.12cm, valign=t]{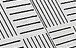} & \includegraphics[width=2.12cm, valign=t]{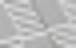} & \includegraphics[width=2.12cm, valign=t]{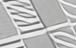} & \includegraphics[width=2.12cm, valign=t]{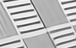} \\
    & HR & LR & SwinIR & MambaIR \\
    & \includegraphics[width=2.12cm, valign=t]{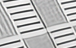} & \includegraphics[width=2.12cm, valign=t]{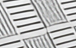} & \includegraphics[width=2.12cm, valign=t]{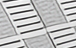} & \includegraphics[width=2.12cm, valign=t]{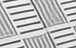} \\
    & CAT-A & ART & MambaIRv2-S & LSM (Ours) \\
    \rule{0pt}{1.5em}
    
    \multirow{4}{*}{\parbox[t]{3cm}{\centering
        \includegraphics[width=2.96cm, valign=t]{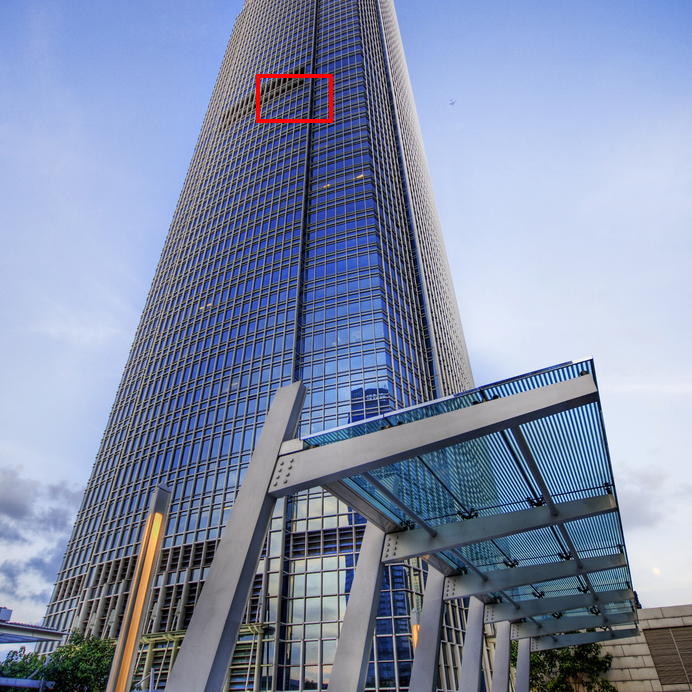}\\
        Urban100: img\_046
    }} & \includegraphics[width=2.12cm, valign=t]{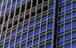} & \includegraphics[width=2.12cm, valign=t]{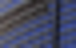} & \includegraphics[width=2.12cm, valign=t]{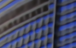} & \includegraphics[width=2.12cm, valign=t]{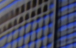} \\
    & HR & LR & SwinIR & MambaIR \\
    & \includegraphics[width=2.12cm, valign=t]{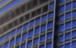} & \includegraphics[width=2.12cm, valign=t]{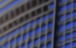} & \includegraphics[width=2.12cm, valign=t]{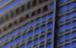} & \includegraphics[width=2.12cm, valign=t]{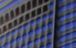} \\
    & CAT-A & ART & MambaIRv2-S & LSM (Ours) \\
    
    \end{tabular}
    }
    \caption{Qualitative comparisons with competitive methods on $\times 4$ classic SR focusing on straight patterns.}
    \label{fig:add_qual1}
\end{figure*}

\begin{figure*}[t]
    \scriptsize
    \renewcommand{\thefigure}{D.4}
    \captionsetup{justification=centering}
    \centering
    \resizebox{\textwidth}{!}{
    \begin{tabular}{ccccc}
    
    \multirow{4}{*}{\parbox[t]{3cm}{\centering
        \includegraphics[width=2.96cm, valign=t]{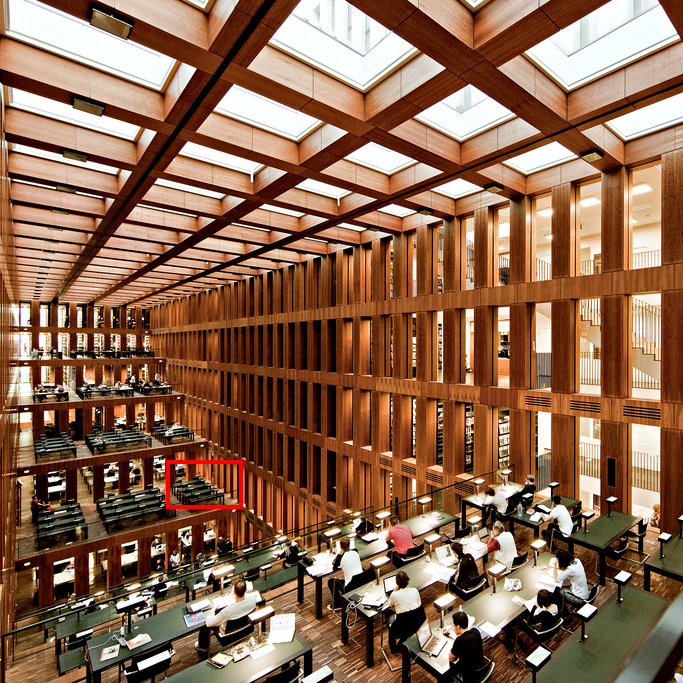}\\
        Urban100: img\_049
    }} & \includegraphics[width=2.12cm, valign=t]{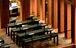} & \includegraphics[width=2.12cm, valign=t]{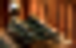} & \includegraphics[width=2.12cm, valign=t]{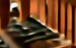} & \includegraphics[width=2.12cm, valign=t]{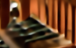} \\
    & HR & LR & SwinIR & MambaIR \\
    & \includegraphics[width=2.12cm, valign=t]{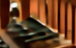} & \includegraphics[width=2.12cm, valign=t]{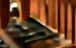} & \includegraphics[width=2.12cm, valign=t]{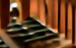} & \includegraphics[width=2.12cm, valign=t]{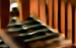} \\
    & CAT-A & ART & MambaIRv2-S & LSM (Ours) \\
    \rule{0pt}{1.5em}
    
    \multirow{4}{*}{\parbox[t]{3cm}{\centering
        \includegraphics[width=2.96cm, valign=t]{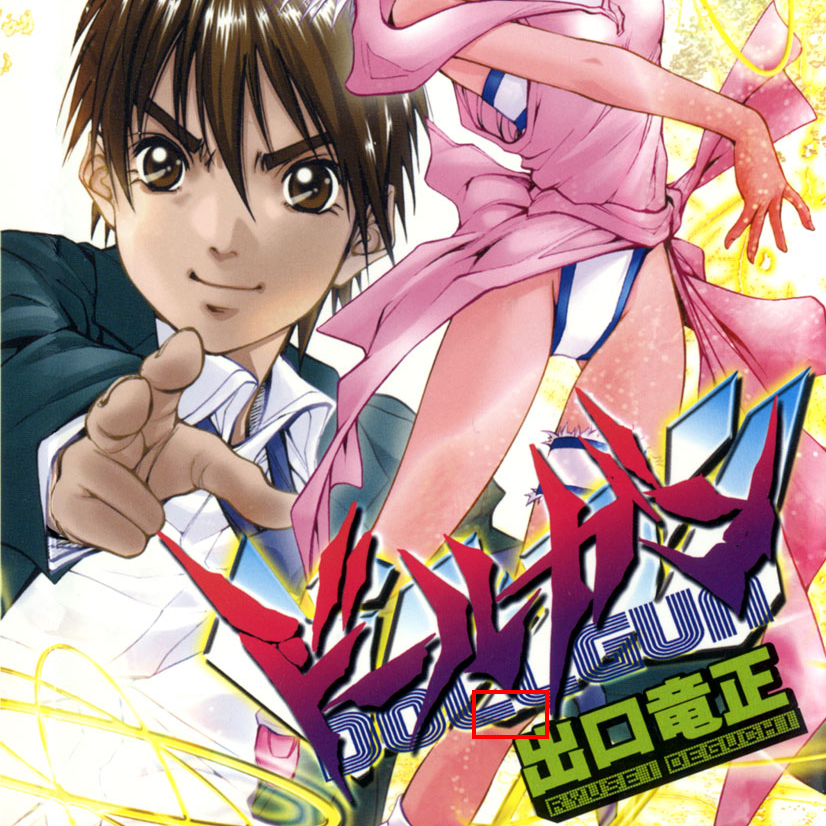}\\
        Manga109: DollGun
    }} & \includegraphics[width=2.12cm, valign=t]{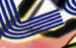} & \includegraphics[width=2.12cm, valign=t]{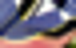} & \includegraphics[width=2.12cm, valign=t]{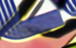} & \includegraphics[width=2.12cm, valign=t]{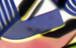} \\
    & HR & LR & SwinIR & MambaIR \\
    & \includegraphics[width=2.12cm, valign=t]{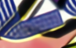} & \includegraphics[width=2.12cm, valign=t]{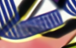} & \includegraphics[width=2.12cm, valign=t]{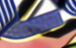} & \includegraphics[width=2.12cm, valign=t]{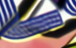} \\
    & CAT-A & ART & MambaIRv2-S & LSM (Ours) \\
    \rule{0pt}{1.5em}
    
    \multirow{4}{*}{\parbox[t]{3cm}{\centering
        \includegraphics[width=2.96cm, valign=t]{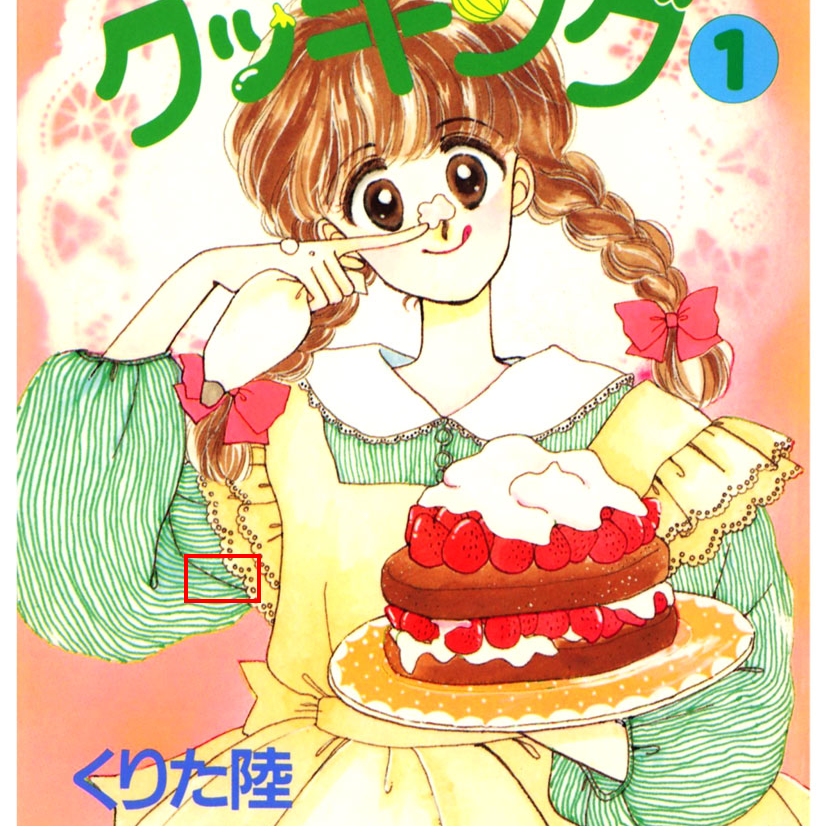}\\
        Manga109: YumeiroCooking
    }} & \includegraphics[width=2.12cm, valign=t]{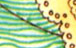} & \includegraphics[width=2.12cm, valign=t]{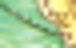} & \includegraphics[width=2.12cm, valign=t]{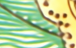} & \includegraphics[width=2.12cm, valign=t]{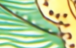} \\
    & HR & LR & SwinIR & MambaIR \\
    & \includegraphics[width=2.12cm, valign=t]{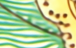} & \includegraphics[width=2.12cm, valign=t]{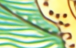} & \includegraphics[width=2.12cm, valign=t]{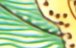} & \includegraphics[width=2.12cm, valign=t]{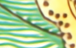} \\
    & CAT-A & ART & MambaIRv2-S & LSM (Ours) \\
    \rule{0pt}{1.5em}
    
    \multirow{4}{*}{\parbox[t]{3cm}{\centering
        \includegraphics[width=2.96cm, valign=t]{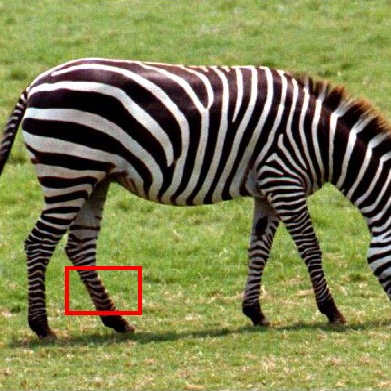}\\
        Set14: zebra
    }} & \includegraphics[width=2.12cm, valign=t]{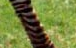} & \includegraphics[width=2.12cm, valign=t]{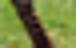} & \includegraphics[width=2.12cm, valign=t]{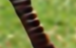} & \includegraphics[width=2.12cm, valign=t]{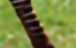} \\
    & HR & LR & SwinIR & MambaIR \\
    & \includegraphics[width=2.12cm, valign=t]{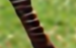} & \includegraphics[width=2.12cm, valign=t]{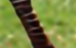} & \includegraphics[width=2.12cm, valign=t]{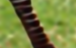} & \includegraphics[width=2.12cm, valign=t]{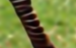} \\
    & CAT-A & ART & MambaIRv2-S & LSM (Ours) \\
    
    \end{tabular}
    }
    \caption{Qualitative comparisons with competitive methods on $\times 4$ classic SR focusing on irregular and curved textures.}
    \label{fig:add_qual2}
\end{figure*}

\noindent \textbf{Qualitative comparison}
Our LSM consistently reconstructs both semantic structures and fine-grained textures across a wide range of images.  
As shown in \cref{fig:add_qual1}, our method recovers structured patterns such as straight lines and architectural details with higher fidelity than competing models on the Urban100 dataset.  
In addition, in \cref{fig:add_qual2}, our approach effectively reconstructs irregular curved textures across datasets while minimizing artifacts that deviate from the ground-truth structure.
\clearpage
\setlength{\tabcolsep}{1pt}
\renewcommand{\arraystretch}{1}
\begin{figure*}[h]
    \footnotesize
    \renewcommand{\thefigure}{D.1}
    \captionsetup{justification=centering}
    \centering
    \begin{tabular}{ccccc}

    \raisebox{0.6in}{\rotatebox{90}{Set5: bird}}
    & \includegraphics[width=4.2cm]{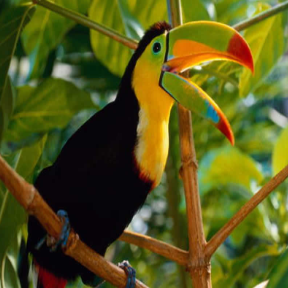}
    & \includegraphics[width=4.2cm]{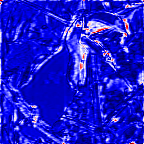}
    & \includegraphics[width=4.2cm]{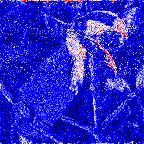}
    & \includegraphics[width=4.2cm]{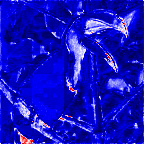}
    \\

     \raisebox{0.13in}{\rotatebox{90}{Manga109: YumeiroCooking}}
    & \includegraphics[trim=0 84 0 280,clip,width=4.2cm]{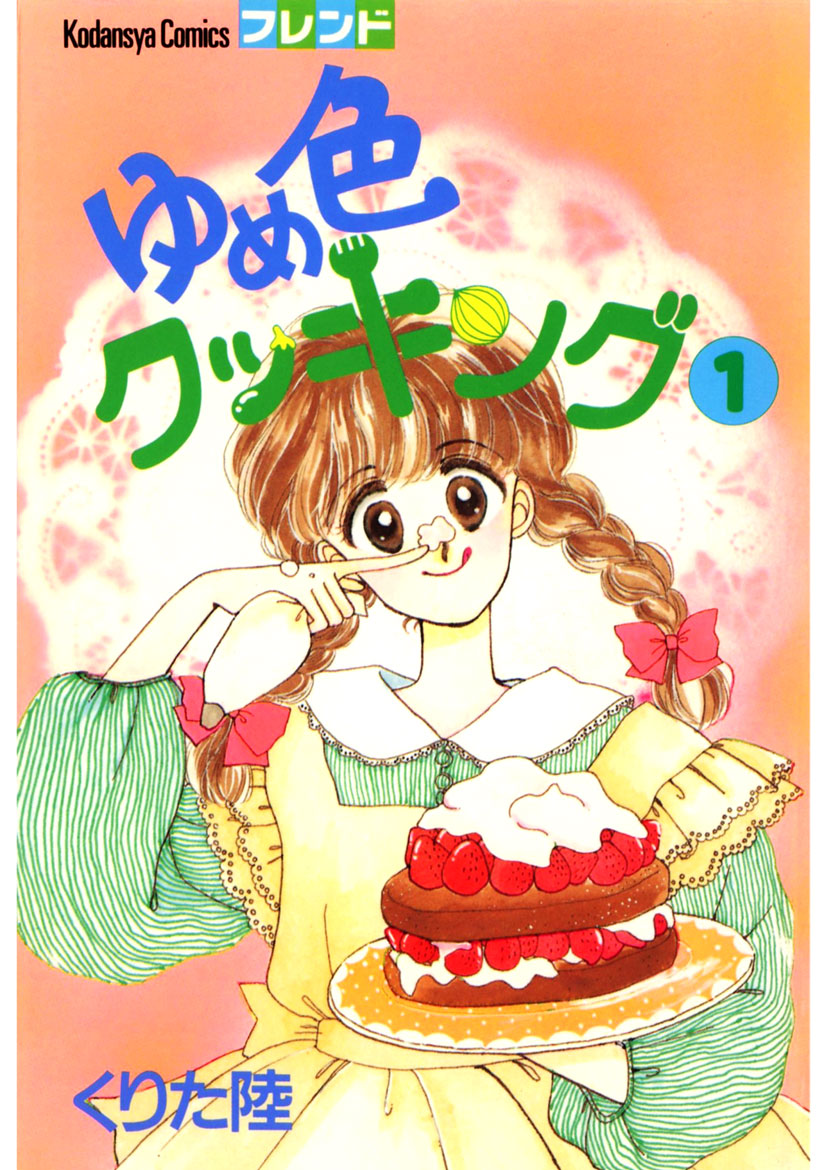}
    & \includegraphics[trim=0 30 0 100,clip,width=4.2cm]{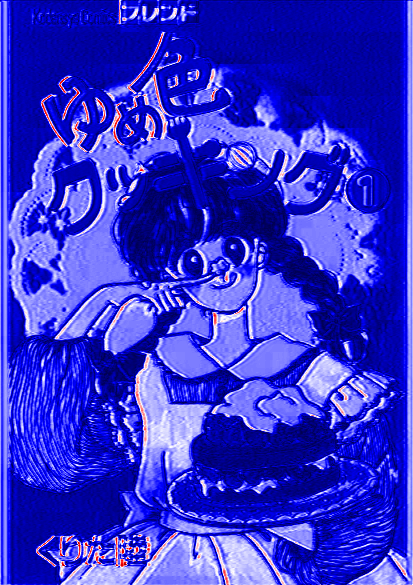}
    & \includegraphics[trim=0 30 0 100,clip,width=4.2cm]{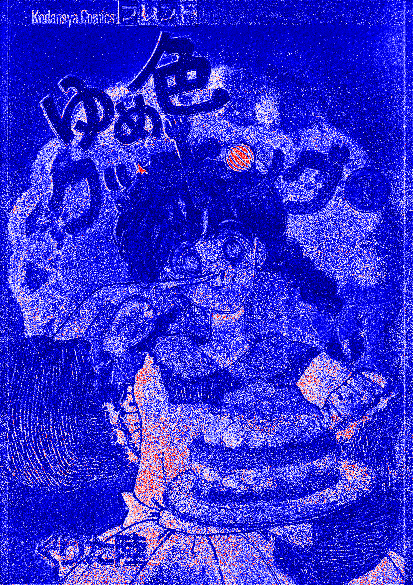}
    & \includegraphics[trim=0 30 0 100,clip,width=4.2cm]{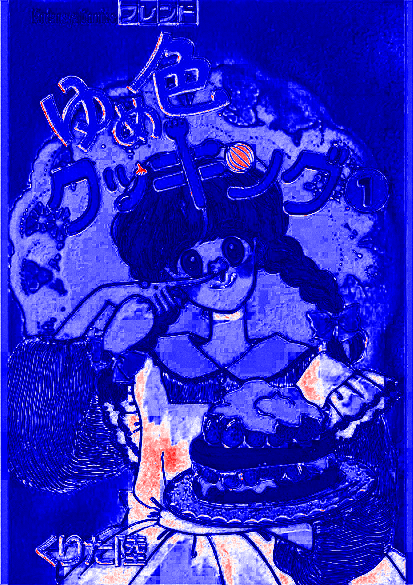}
    \\

    \raisebox{0.155in}{\rotatebox{90}{Urban100: img\_012}}
    & \includegraphics[width=4.2cm]{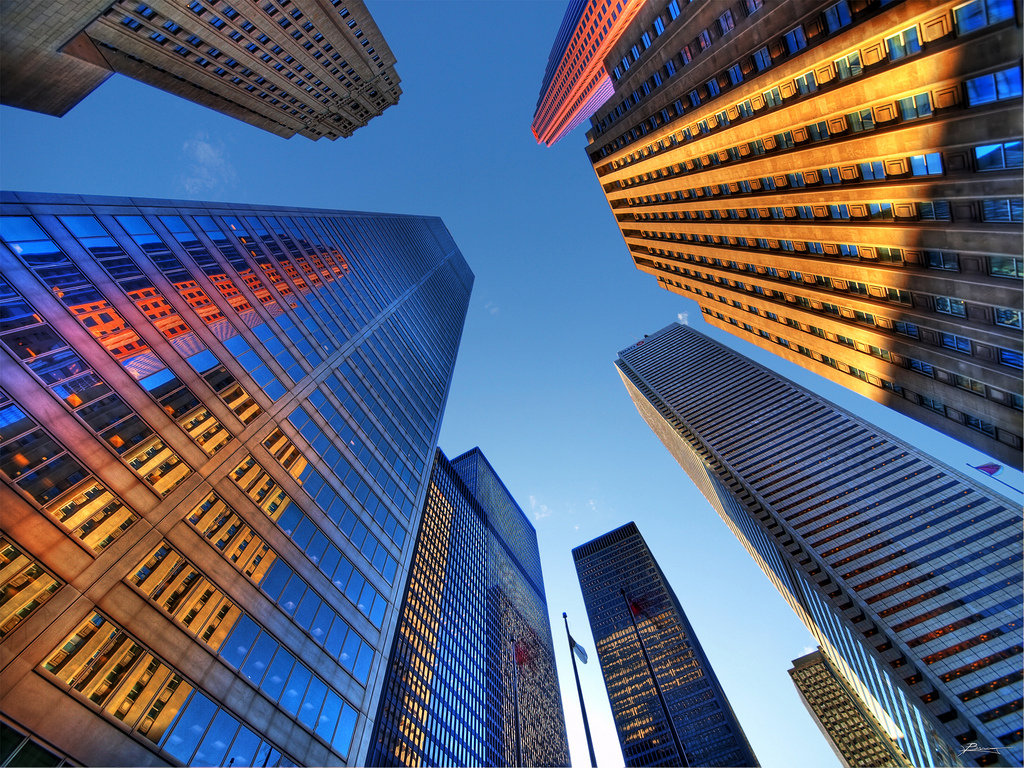}
    & \includegraphics[width=4.2cm]{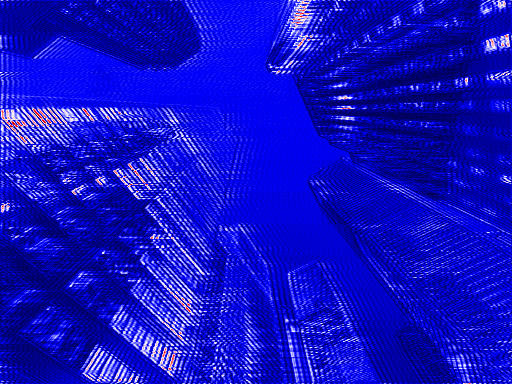}
    & \includegraphics[width=4.2cm]{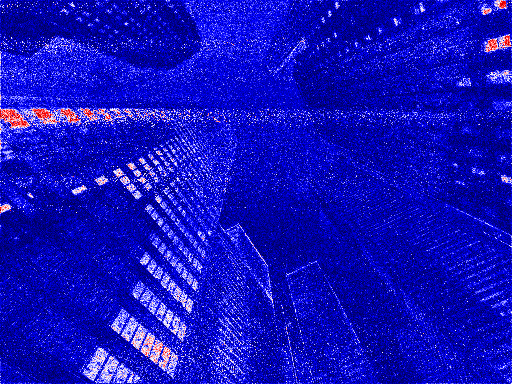}
    & \includegraphics[width=4.2cm]{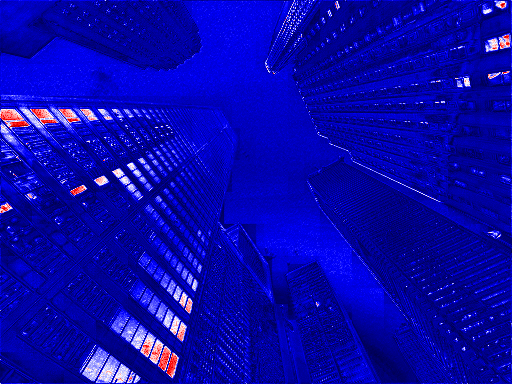}
    \\

    \raisebox{0.155in}{\rotatebox{90}{Urban100: img\_070}}
    & \includegraphics[width=4.2cm]{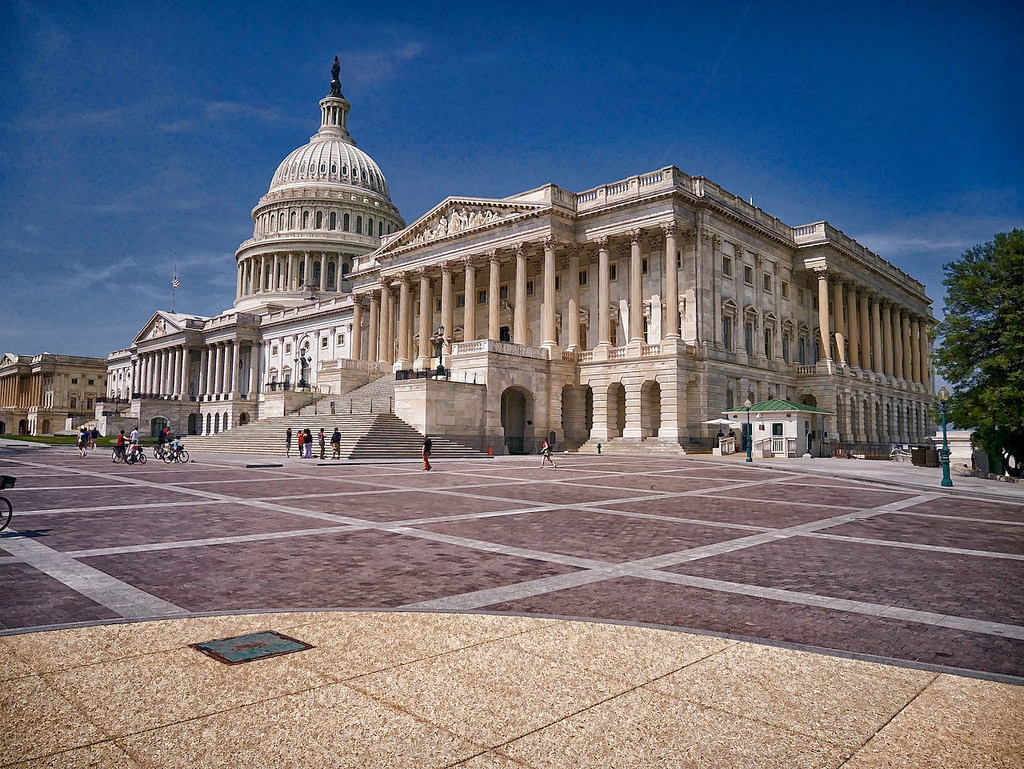}
    & \includegraphics[width=4.2cm]{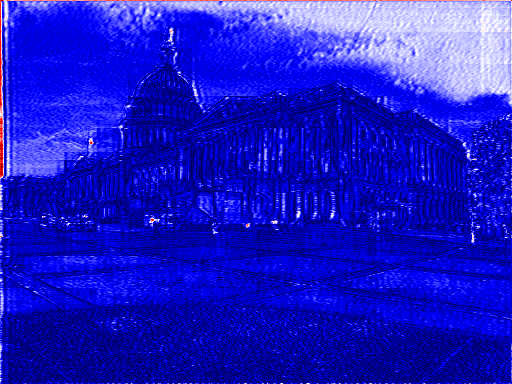}
    & \includegraphics[width=4.2cm]{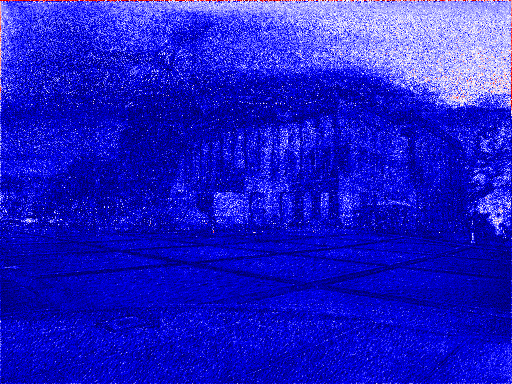}
    & \includegraphics[width=4.2cm]{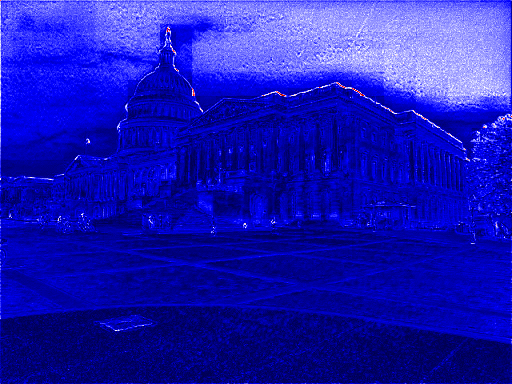}
    \\
    \vspace{-0.5pt}
    & \multicolumn{1}{c}{HR} & \multicolumn{1}{c}{Vanilla LRU} & \multicolumn{1}{c}{LRU + Categorize} & \multicolumn{1}{c}{LRU + Categorize \& Modulate} \\
    
    \end{tabular}
    \vspace{-0pt}
    \caption{Visualization of hidden states.}
    \label{fig:add_visual}
    \vspace{-50pt}
    \addtocounter{figure}{-1}
\end{figure*}
\setlength{\tabcolsep}{1pt}
\renewcommand{\arraystretch}{1}
\begin{figure*}[h]
    \footnotesize
    \renewcommand{\thefigure}{D.2}
    \captionsetup{justification=centering}
    \centering
    \begin{tabular}{ccccc}
    \raisebox{0.10in}{\rotatebox{90}{Urban100: img\_098}}
    & \includegraphics[trim=0 0 0 10,clip,width=4.2cm]{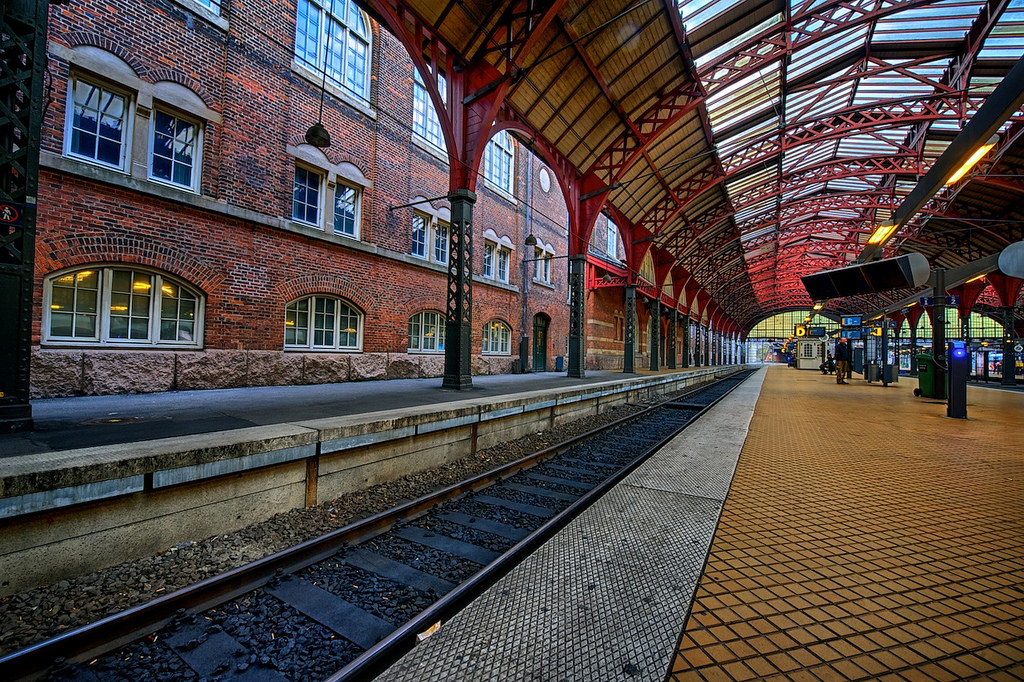}
    & \includegraphics[trim=12 12 12 12,clip,width=4.2cm]{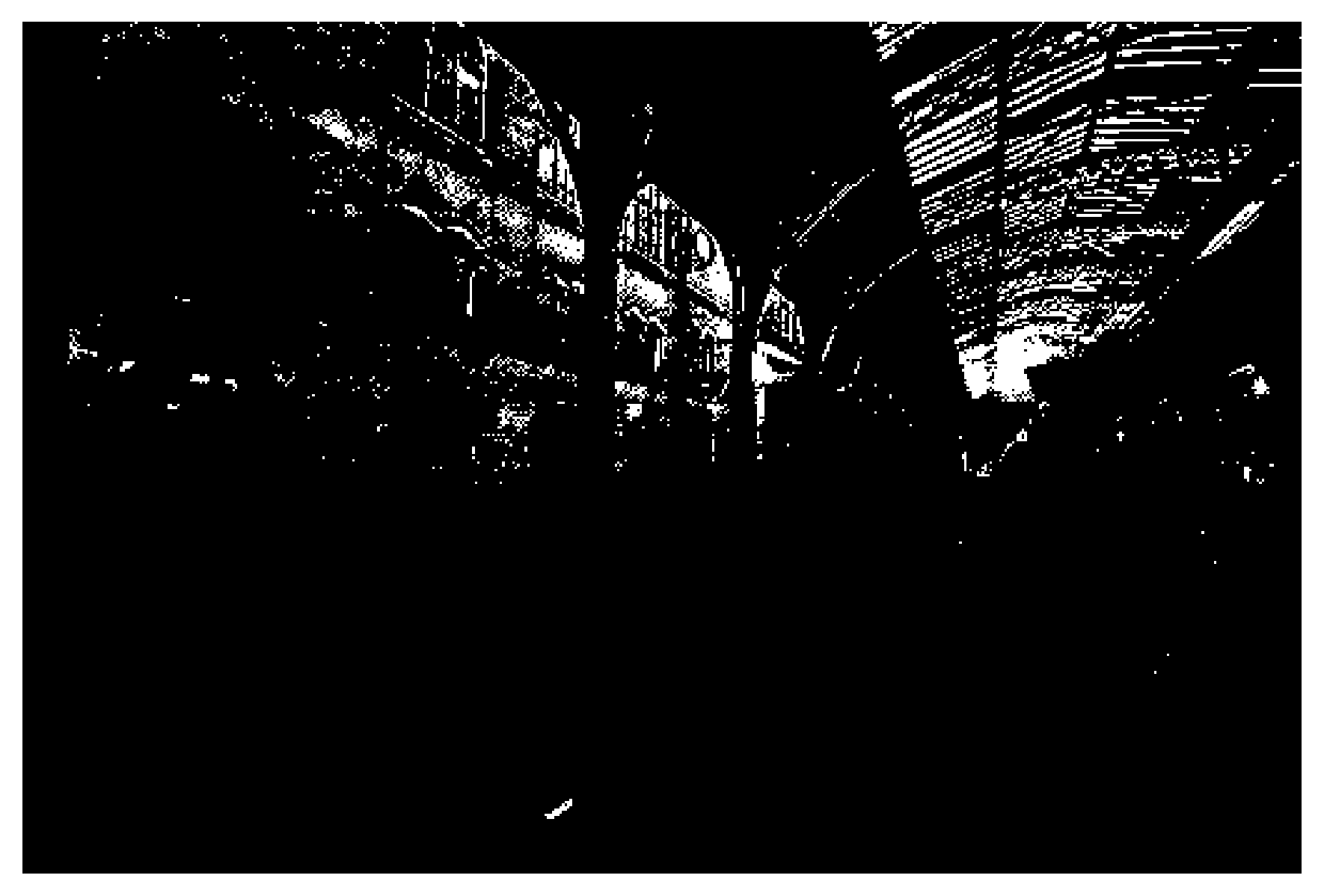}
    & \includegraphics[trim=12 12 12 12,clip,width=4.2cm]{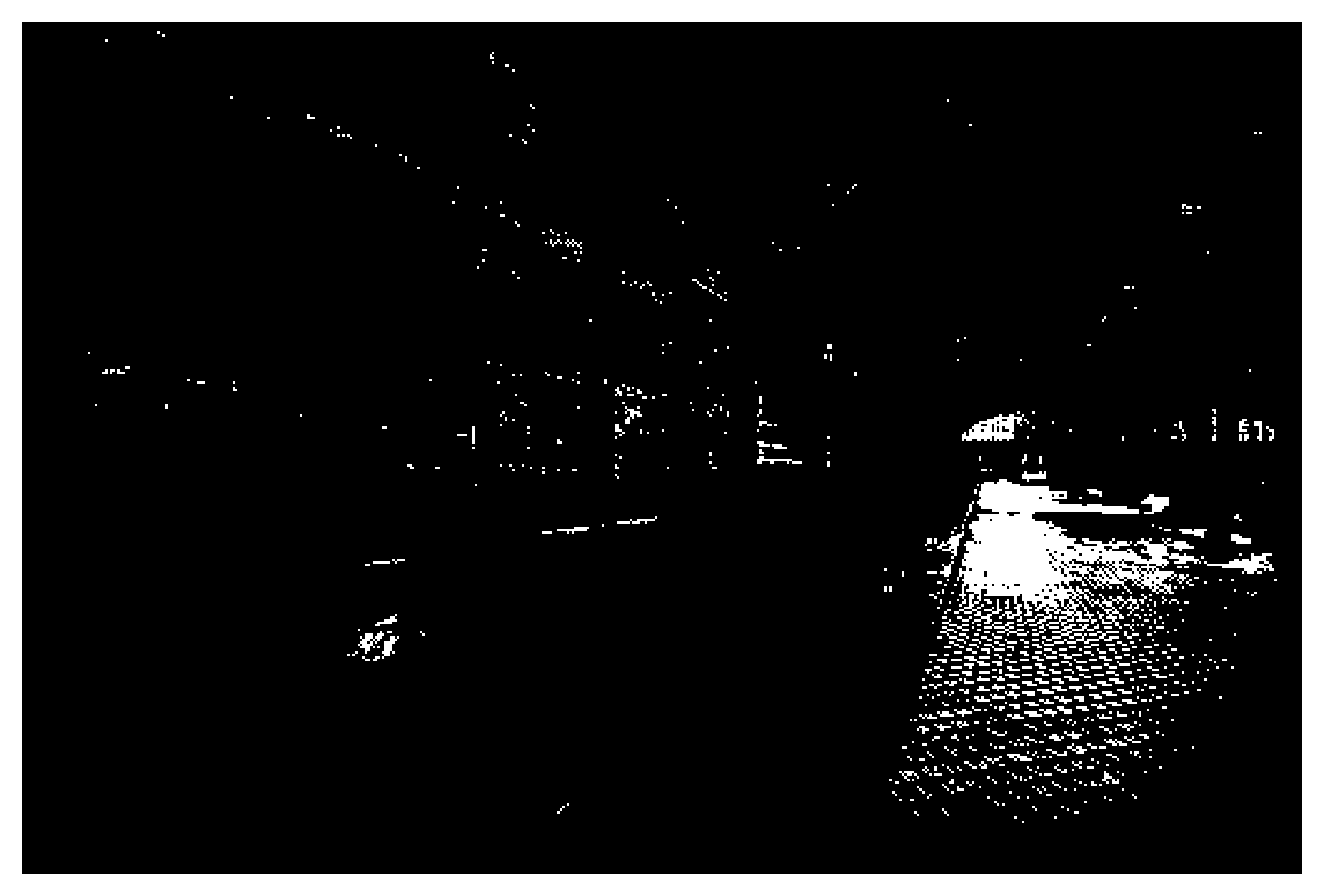}
    & \includegraphics[trim=12 12 12 12,clip,width=4.2cm]{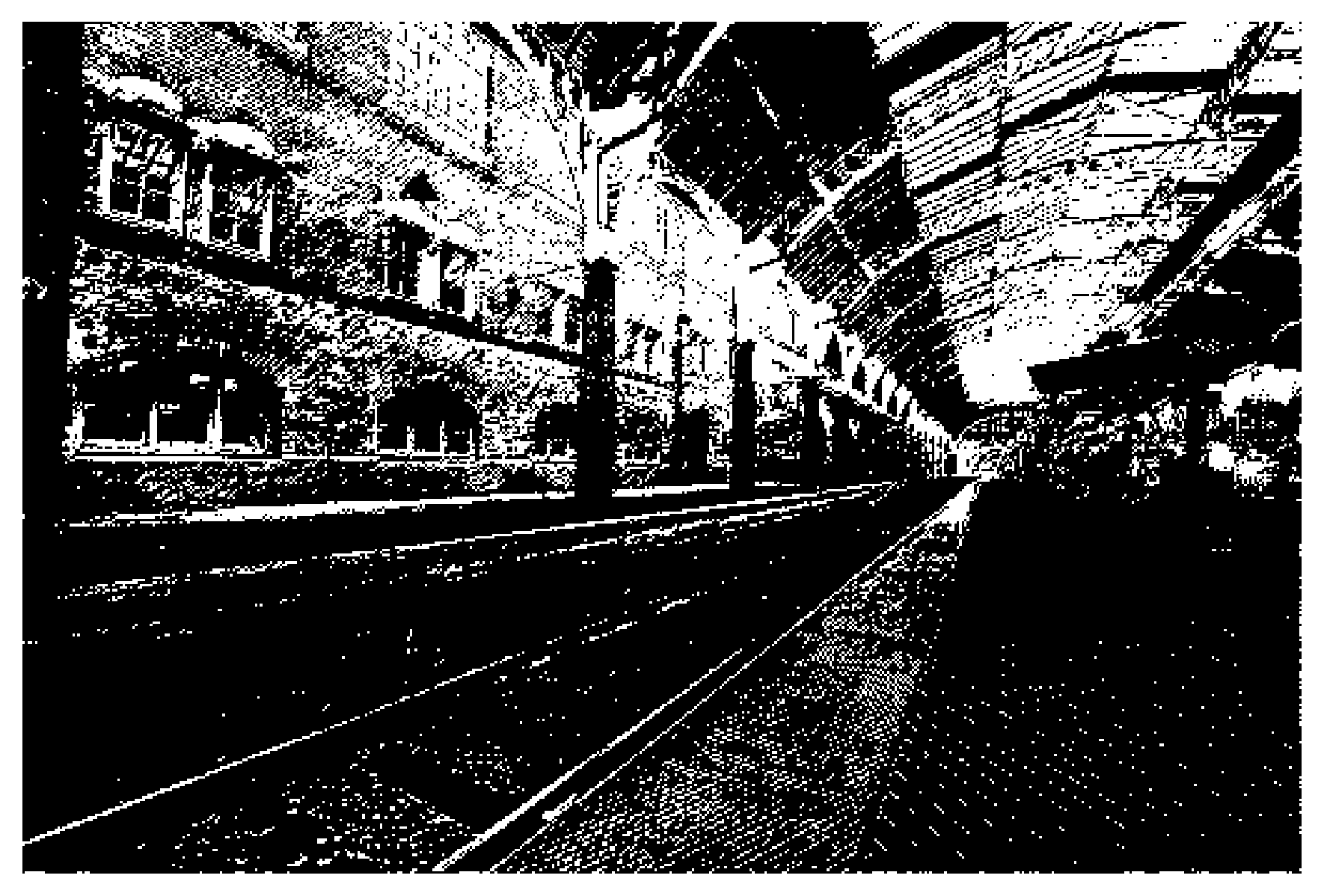}
    \\
    \vspace{-0.5pt}
    & \multicolumn{1}{c}{HR} & \multicolumn{1}{c}{Category 1} & \multicolumn{1}{c}{Category 2} & \multicolumn{1}{c}{Category 3} \\
    
    \end{tabular}
    \vspace{-0pt}
    \caption{Visualization of categorization results.}
    \label{fig:add_category}
    \vspace{-0.3cm}
    \addtocounter{figure}{-1}
\end{figure*}
\clearpage
\setlength{\tabcolsep}{1pt}
\renewcommand{\arraystretch}{1} 

\clearpage
\renewcommand{\arraystretch}{1}

\end{document}